%% file: main.tex
\documentclass[conference]{IEEEtran}
\usepackage{graphicx} 
\usepackage{color,soul}
\usepackage{paralist}
\usepackage{enumitem}
\usepackage{url}
\usepackage{amsfonts}
\usepackage{amssymb}
\usepackage{wrapfig}
\usepackage{amsmath}
\usepackage{subfigure}
\usepackage{gensymb}
\usepackage{multirow}
\usepackage{algorithm}
\usepackage{algorithmicx}
\usepackage{algpseudocode}
\usepackage{comment}
\usepackage{balance}
\usepackage[mathscr]{euscript}
\usepackage{caption}
\usepackage{float}
\usepackage{amsbsy}
\usepackage{bm}
\usepackage{amsthm}
\usepackage{amssymb}
\usepackage{diagbox}
\newcommand{\ignore}[1]{}

\theoremstyle{definition}

\renewcommand{\algorithmicrequire}{ \textbf{Input:}}

\begin{document}

\pagestyle{plain}

\title{A Comprehensive and Reliable Feature Attribution Method: Double-sided Remove and Reconstruct (DoRaR)}

\author{
 \IEEEauthorblockN{Dong Qin, George Amariucai*, Daji Qiao, Yong Guan, Shen Fu}
 \IEEEauthorblockA{
 {Iowa State University, 
 Ames, IA, USA},
 \{dqin, daji, guan, shenfu\}@iastate.edu}
 \IEEEauthorblockA{{*Kansas State University, 
 Manhattan, KS, USA},
 amariucai@ksu.edu}\\
}

\maketitle
\input{paper_body}

\balance

\bibliographystyle{IEEEtran}

\appendices
\input{Supplementary_Materials}
\end{document}

%% file: paper_body.tex
\begin{abstract}
The limited transparency of the inner decision-making mechanism in deep neural networks (DNN) and other machine learning (ML) models has hindered their application in several domains. In order to tackle this issue, feature attribution methods have been developed to identify the crucial features that heavily influence decisions made by these black box models. However, many feature attribution methods have inherent downsides. For example, one category of feature attribution methods suffers from the artifacts problem, which feeds out-of-distribution masked inputs directly through the classifier that was originally trained on natural data points. Another category of feature attribution method finds explanations by using jointly trained feature selectors and predictors. While avoiding the artifacts problem, this new category suffers from the Encoding Prediction in the Explanation (EPITE) problem, in which the predictor's decisions rely not on the features, but on the masks that selects those features. As a result, the credibility of attribution results is undermined by these downsides. In this research, we introduce the Double-sided Remove and Reconstruct (DoRaR) feature attribution method based on several improvement methods that addresses these issues. By conducting thorough testing on MNIST, CIFAR10 and our own synthetic dataset, we demonstrate that the DoRaR feature attribution method can effectively bypass the above issues and can aid in training a feature selector that outperforms other state-of-the-art feature attribution methods. Our code is available at https://github.com/dxq21/DoRaR.
\end{abstract}

\input{introduction.tex}

\input{problem_statement}

\input{algorithm.tex}

\input{evaluation_method}

\input{Comparison_to_other_methods}

\input{discussion.tex}

\input{related_work}

%% file: introduction.tex
\section{Introduction}
\label{sec:introduction}
\subsection{Background and Motivation}

The machine learning models are spreading at a high speed in many crucial aspects of society, powered by complex models such as deep neural networks (DNN). These models have a wide application in the real world and such a trend has made interpretable machine learning consequential for trusting model decisions  \cite{lipton2018mythos} and expanding knowledge  \cite{silver2017mastering}. Interpretability in machine learning is well-studied in order to make machine learning models better serve human beings. On the one hand, efforts have been made to illuminate the inner mechanism of some deep learning models, making them more transparent, e.g. global self interpretable models or global explanations  \cite{du2019techniques}. On the other hand, many other methods have been presented to provide an understanding of which features locally, for a given instance of data, are more important than others in final decision making. This type of explanation can be categorised as feature attribution method. For example, feature attribution method produces masks to explain images, where important pixels are highlighted based on their contribution to the target label prediction. The first approach helps people trust a model and the second approach lets users trust the prediction result. Our research focuses on feature attribution method that make the prediction trustworthy for users.

Providing a trustworthy DNN model explanation efficiently is challenging due to limitations in many aspects. For example, perturbation based methods \cite{zeiler2014visualizing, zhou2015predicting, zintgraf2017visualizing, fong2017interpretable} and locally linear methods \cite{dabkowski2017real, ribeiro2016should, lundberg2017unified} are computationally inefficient, which makes them inapplicable in the industry. Other methods like gradient based methods are inaccurate and vulnerable to adversarial manipulation  \cite{dombrowski2019explanations,heo2019fooling}.

In recent years, some studies such as \cite{dabkowski2017real, chen2018learning, yoon2018invase, fu2021differentiated}, generate a mask to select features and then feed the masked input to the classifier that was trained on the complete feature dataset. While this method is efficient, it faces the issue of unwanted \textbf{artifacts} \cite{dabkowski2017real}, because the masked inputs typically fall outside the natural distribution of the training dataset. Further details regarding the artifacts problem are discussed in Section \ref{subsec:Artifacts}.

In order to solve the artifacts problem, some efforts have been made, e.g., inducing $\alpha$-norm \cite{mahendran2015understanding}, Gaussian blur \cite{yosinski2015understanding} or utilizing low resolution intermediate activation features in the CNN model, followed by up-sampling \cite{du2018towards}. These approaches mitigate the problem but do not solve it essentially, and may bring in other side-effects, e.g., introducing extra evidence, reducing prediction accuracy and so on.

Inspired by \cite{hooker2018benchmark}, some methods \cite{bang2021explaining,chen2018learning,yoon2018invase} retrain a new classifier with masked input, and then evaluate explanations by their performance in the new classifier. However, training a model based on masked inputs could cause a new problem which we call Encoding Prediction in the Explanation (EPITE) -- details of the EPITE problem are described in Section~\ref{subsec:EPITE}. This problem is commonly seen in feature attribution methods with an encoder-decoder joint training architecture.

In this paper, we propose a reliable and comprehensive feature attribution method, which we call Double-sided Remove and Reconstruct (DoRaR), to interpret a neural network classifier. In this model, we try to explain how a pre-trained classifier works by finding the most contributing explanation units in the input sample. The number and size of such units (e.g. 4 4$\times$ 4 pixels based chunk) are predefined. A feature selector is trained to find these units. Selected features will be used to train a generative model to reconstruct a sample. Then, instead of the masked sample, our reconstructed sample will be used to predict the target label in the pre-trained classifier. The non-selected part is treated similarly, i.e., it is used to train another generative model to reconstruct an sample, which is then evaluated by the pre-trained classifier. Therefore, both selected features and non-selected features are evaluated and their prediction losses are used to train the feature selector.

\subsection{Contribution}
The contributions of this research are as follows:

1) We present a feature attribution method (DoRaR) which can deal with both the artifacts and the EPITE problems.

2) We present a new definition of feature selector which defines both size and number of required explanation units. It limits the feature interaction within a clearly defined scenario.

3) Our feature attribution method is compared with other state-of-the-art feature attribution methods including LIME, Grad, SmoothGrad, VIBI, Real-X and Guided Feature Inversion, \cite{ribeiro2016should,simonyan2013deep,smilkov2017smoothgrad,bang2021explaining,jethani2021have,du2018towards} on MNIST, CIFAR-10 and a user-mouse interaction behavioural based datasets using an appropriate testing scheme.

%% file: problem_statement.tex
\section{Problem Statement}

\begin{figure}[htb]
\centering\includegraphics[width=1\columnwidth]{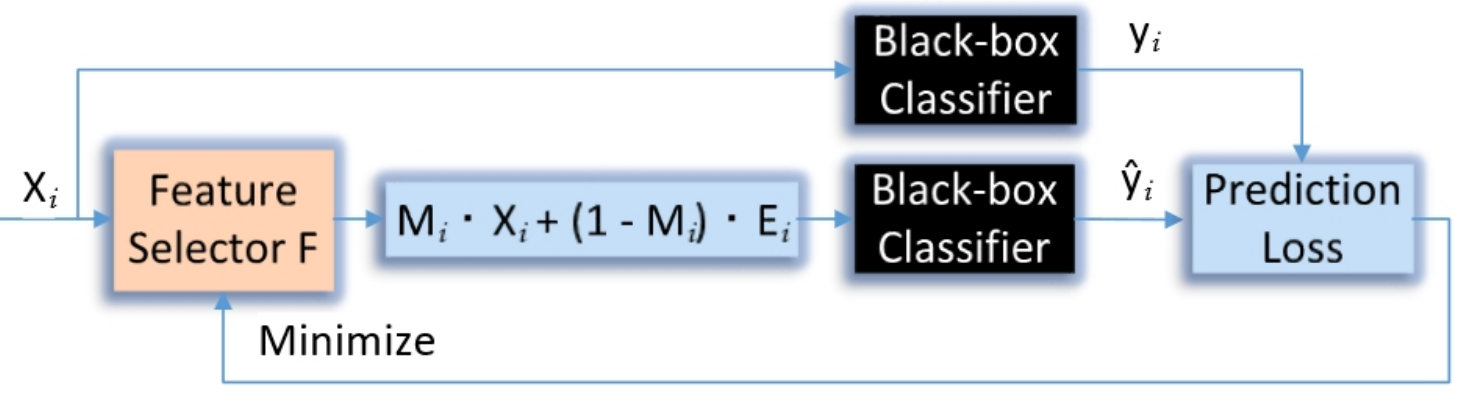}

\caption{Feature attribution method explains a black-box classifier by generating a relevancy score vector $M_i$ that highlight those key features in the input. $E_i$ represents the background noise used to fill the non-selected features.}
\label{fig:FAM with artifacts}
\end{figure}

In this section, we define the goal of interpreting a DNN-based classifier and describe possible problems occurring in some feature attribution methods. In a multi-class classification problem, the classifier can be defined as $\bm{y}_i = P(\bm{X}_i)$ as shown in Fig.\ref{fig:FAM with artifacts}. We aim to find the contributing factors, which we called explanation, in $\bm{X}_i$ that lead the black-box classifier $P$ to make the prediction $\bm{y}_i$. More specifically, given an input $\bm{X}_i$, we want to find a discrete mask $\bm{M}_i^*$, where $\bm{M}_i^*(j)\in\{0,1\}$ represents the corresponding value at dimension $j$, that selects the features of $\bm{X}_i$ with the highest relevance for explaining $\bm{y}_i = P(\bm{X}_i)$.

To achieve this goal, some algorithms have been presented in previous research \cite{fong2017interpretable,dabkowski2017real,chen2018learning,yoon2018invase, du2018towards}. As shown in Fig.~\ref{fig:FAM with artifacts}, a typical category is the non-retraining based algorithm which evaluates the contribution of selected key features in the pre-trained black-box classifier. This category of feature attribution method always suffer from the \textbf{1) Artifacts problem}, because the masked input sample is out of natural data distribution which the classifier is trained with.

\begin{figure}[htb]
\centering\includegraphics[width=1\columnwidth]{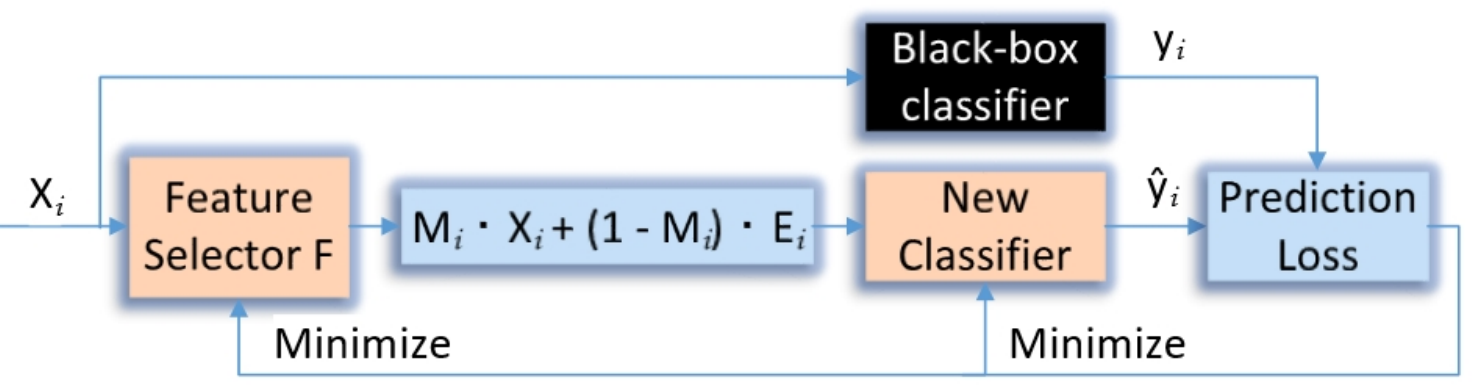}
\caption{Feature attribution method trains a feature selector and a new classifier to replicate the black-box classifier performance and catering for new data distribution.}
\label{fig:FAM with EPITE}
\end{figure}

In order to address the artifacts problem, re-training a new classifier to evaluate the selected key features is introduced in \cite{hooker2018benchmark}. Inspired by the re-training strategy, multiple feature attribution methods has been presented to find the key features \cite{fu2021differentiated,bang2021explaining,jethani2021have}. However, this category of feature attribution method may suffer from the \textbf{2) EPITE problem}, because the class information might be leaked through the mask \cite{rong2022consistent}. The final prediction can be made by the re-trained classifier which learns the class label only by the mask shape rather than the feature values selected by the mask.

These problems jeopardizes their model explanation's trustworthiness. Current feature attribution methods either don't address the artifacts problem at all \cite{dabkowski2017real,fong2017interpretable,du2018towards,fu2021differentiated} or address it but introduce the EPITE problem \cite{bang2021explaining,chen2018learning,yoon2018invase,hooker2018benchmark}. None of them essentially solves all problems at the same time.

In this research, we introduce a new feature attribution method for training a feature selector, both the artifacts problem and the EPITE problem can be avoided in our feature attribution method.

 \subsection{The Artifacts Problem}
 \label{subsec:Artifacts}

\begin{figure}[htb]

\centering\includegraphics[width=0.5\columnwidth]{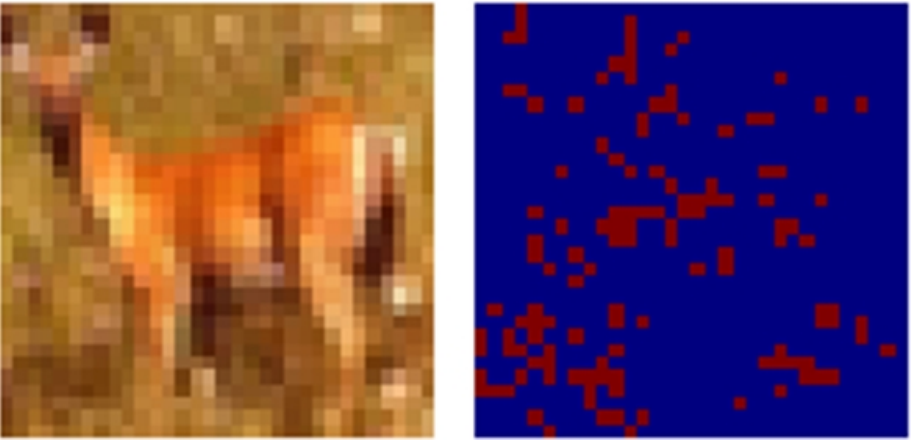}

\caption{Pixel based explanation generated by gradient based feature attribution method, red pixels represent those features that are more important for classification. Some pixels from irrelevant area are selected as explanations due to the artifacts problem.}
\label{fig:artifacts example}
\end{figure}

\begin{figure}[htb]
\centering
    \subfigure[A classifier learned from the data of two classes, the green part is more determined by the natural data distribution, and the orange part is more determined by random factors like randomly initialized parameters in the neural network classifier.]{\label{fig:art1}\includegraphics[width=0.45\columnwidth]{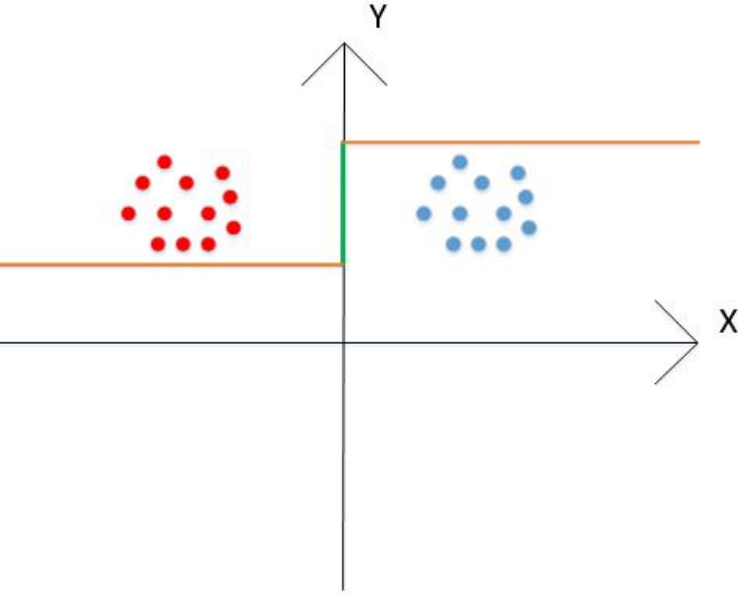}}
    \hspace{3mm}
\centering
    \subfigure[The x-axis feature is maintained, while the y-axis feature is removed and replaced by 0, which leads to wrong classifications of red data points. The sever accuracy drop will make y-axis feature been regarded as the key feature.]{\label{fig:art2}\includegraphics[width=0.45\columnwidth]{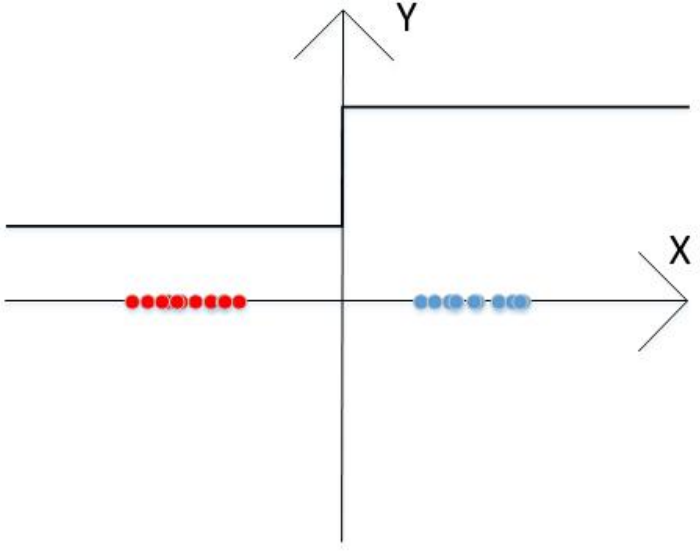}}
\centering
\caption{An example of a neural network classifier learned from two dimensional data of two classes. We try to find the key feature that is more relevant to the correct classification by perturbing one feature.}

\label{fig:artifacts}
\end{figure}

Neural networks are known to be affected by surprising artifacts \cite{mahendran2015understanding,nguyen2015deep}. Fig.~\ref{fig:artifacts example} shows an example of the artifacts problem, where some obviously irrelevant pixels are selected as key features. Fig.~\ref{fig:art1} illustrates how the artifacts problem happens, a classifier represented here as the green and orange lines is trained on a dataset, shown as red and blue data points, with two features x and y. The green part of the classifier boundary is mainly shaped by the nature of the data distribution. However, the orange part of the classifier boundary is somewhat arbitrary, and may depend on randomly initialized neural network parameters, optimizer settings and so on. When evaluating an explanation obtained by replacing part of the features by ad-hoc values, such as zeros -- such an explanation, which drops the $y$ feature and replaces it with zeros as shown in Fig.~\ref{fig:art2} -- the data points follow a distribution different than the original natural distribution, and they may be misclassified by the classifier. The severe decrease in classification accuracy will in this case lead to the wrong conclusion that the y-axis feature is the key factor for maintaining correct classification, so it should be chosen as the explanation. In this case, setting the y-axis feature to zero creates the unwanted artifact in the explanation that we are trying to avoid.

 More generally, for higher dimensional cases such as images and for different types of background, such as mean value, Gaussian blurred noise and so on, it is possible to find certain masks $\bm{M}_i$ (artifacts) for input $\bm{X}_i$ which make the image follow a different distribution than the natural one, and generate unexpected output \cite{dabkowski2017real}. Therefore, these artifacts selected by $\bm{M}_i$ could become part of our explanation, but such explanations may not be meaningful and should be avoided.

 \subsection{The EPITE Problem}
 \label{subsec:EPITE}

 The EPITE problem commonly occurs in retrain evaluation strategy \cite{hooker2018benchmark} or feature attribution methods that rely on a retrained predictor to evaluate their feature attribution result \cite{bang2021explaining,chen2018learning,yoon2018invase}. These methods typically train a feature selector to perform instance-wise feature selection and feed the selected features to a predictor, which learns to predict the target label based on the selected features. Such encoder-decoder structure trains the feature selector and the predictor jointly to optimize the overall prediction accuracy.

\begin{figure}[htb]
\centering\includegraphics[width=0.75\columnwidth]{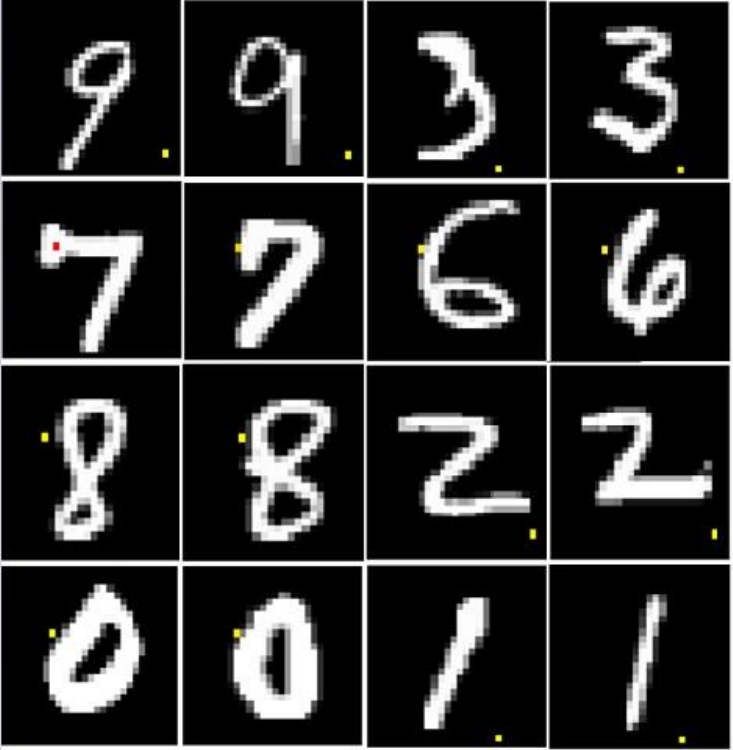}
\caption{With a retrained predictor, a single pixel (highlighted in yellow) based explanation achieves in high prediction accuracy.}
\label{fig:EPITE}
\end{figure}

 However, while this evaluation strategy overcome the artifacts problem, it comes with a problem that the encoder part of the model could encode the prediction label in the explanation. Fig.~\ref{fig:EPITE} shows an example of the EPITE problem. In this example, an encoder-decoder structure based feature attribution method VIBI \cite{bang2021explaining} achieves high prediction accuracy of 72\% in the retrained predictor with a single selected pixel from irrelevant areas as input and the rest part replaced by 0. This is achievable because the feature selector encodes the label information within the index of the selected feature rather than its value.

The EPITE problem is first being mentioned by \cite{jethani2021have}, where an algorithm called REAL-X is introduced to solve it. The claim is that the EPITE problem is solved by first training a predictor, called Eval-X, using randomly chosen features, then training the Real-X feature selector through the feedback from the predictor. But the authors admit it is difficult to reach optimality in practice by training the Eval-X through randomly chosen features, especially when the feature dimension is high. Besides, their feature selector is trained based on their Eval-X predictor, so their feature selctor Real-X is learned to find key features only for the Eval-X predictor rather than arbitrary machine learning predictors.

In \cite{rong2022consistent}, the EPITE problem is referred to as the class information leakage through masking. In such case, the retrained predictor can predict the label purely by indexes of the selected features, in other words the mask shape, rather than values of the selected features. To address the EPITE problem, the authors propose a solution that involves filling the removed missing features with weighted sums of adjacent selected features as background $\bm{E}$. This minimizes $I(\bm{X'};\bm{M})$, where $\bm{X'}=\bm{X}\cdot \bm{M}+\bm{E}\cdot(1-\bm{M})$, making it harder for the retrained model to recognize the shape of the mask and make corresponding predictions. However, this method is not applicable when the selected features are very few.

%% file: algorithm.tex
\section{The Proposed Double-sided Remove and Reconstruct (DoRaR) Solution}
\label{sec:DoRaR}
In this paper, we try to overcome the problems we may face in a feature attribution method. One is the EPITE problem, very common in the encoder-decoder joint training structure, and other is the artifacts problem, which often emerges when evaluating masked input on a classifier trained with dataset that follows the natural data distribution.

\subsection{The DoRaR Feature Attribution Method}

\begin{table}[h]
\centering
\caption{Notation. We use capital script letters for sets, capital bold letters for multi-dimensional arrays (like images), lowercase bold letters for single-dimensional arrays (like labels), regular lowercase letters for scalars, and regular capital letters for functions.}
\label{table:notation}
\scalebox{1}{
\begin{tabular}{|l|l|}
\hline

$\pmb{\mathscr{X}}$              & Training set                                     \\ \hline
$\bm{X}_i$              & $i$th sample of $\pmb{\mathscr{X}}$                            \\ \hline
s                           & Size of the training set                     \\ \hline
$m$              & Feature size of samples in dataset $\pmb{\mathscr{X}}$                     \\ \hline
$\bm{X}_i(j)$                    & $j$th feature of $\bm{X}_i$                              \\ \hline
$\bm{E}_i(j)$                    & $j$th feature of the $i$th sample's background noise \\ \hline

$P$                    & Pre-trained classifier  \\ \hline
$F$                    & Feature selector  \\ \hline
$\bm{\theta}$                    & Model parameters  \\ \hline
$n_e$, $s_e$                    & \begin{tabular}[c]{@{}l@{}}Number and size of output explanation units for feature\\ selector\end{tabular} \\ \hline
$G$                    & Generative model  \\ \hline
$\widehat{\bm{X}_i}$, $\Tilde{\bm{X}_i}$        & Reconstructed samples from selected and rest features \\ \hline
$\bm{y}, \widehat{\bm{y}}$        & Prediction label of original and reconstructed sample \\ \hline
$H_{\bm{y}}(\widehat{\bm{y}})$        & Binary Cross Entropy of $\bm{y}$ and $\widehat{\bm{y}}$ \\ \hline
$\bm{M}_i$                           & Soft mask                                                    \\ \hline
$\bm{M}_i^*$                           & Discrete mask                                                    \\ \hline
d                           & Batch size of model training                      \\ \hline
$l_1$, $l_2$           & \begin{tabular}[c]{@{}l@{}}Reconstruction loss for the selected features and the\\ complementary part\end{tabular} \\ \hline
$l_3$, $l_4$        & \begin{tabular}[c]{@{}l@{}}Prediction loss for the selected features and the\\ complementary part\end{tabular} \\ \hline
$\alpha$, $\beta$    & Control parameters for balancing $l_1 \sim l_4$ \\ \hline
$a_1$                           & Prediction accuracy achieved with selected features                      \\ \hline
$a_2$                           & Prediction accuracy achieved with non-selected features                      \\ \hline
\end{tabular}
}
\end{table}

 We combine multiple improvement techniques and propose our DoRaR feature attribution method. Table.~\ref{table:notation} summarizes the notations used in following sections.

\begin{figure*}[htb]

\centering\includegraphics[width=1\linewidth]{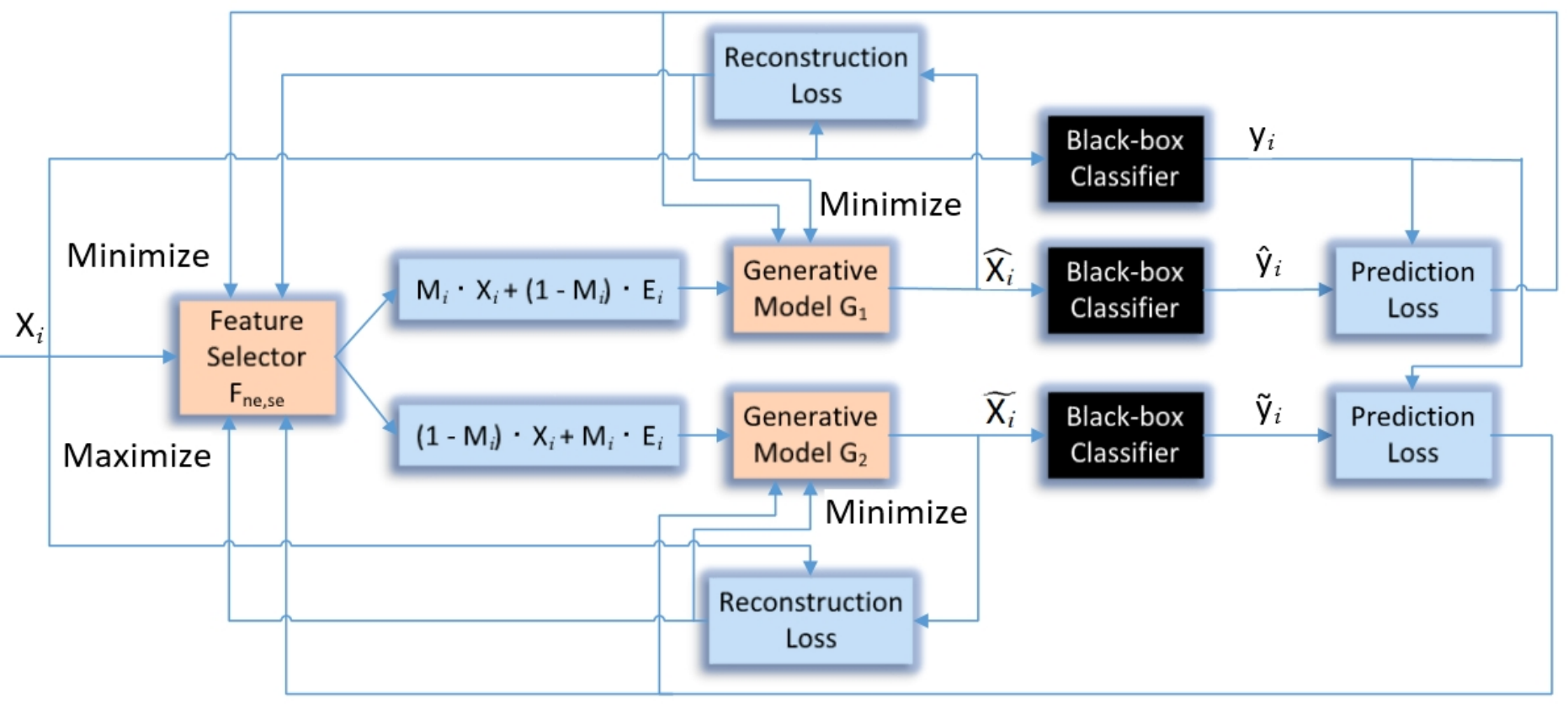}
\caption{DoRaR feature selector training diagram. Feature selector $F_{n_e,s_e}$ that selects $n_e$ explanation units with size $s_e$(e.g. 4 chunks of 4$\times$ 4 pixels, 1 sequence of 16 points) is trained to minimize prediction and reconstruction losses for selected features and maximize prediction and reconstruction losses for non-selected features. Generative models $G_1$ and $G_2$ are trained in the same time as the feature selector to minimize prediction and reconstruction losses of corresponding selected and non-selected features. $E_i$ refer to the background noise randomly draw from empirical data distribution to fill the missing area.}
\vspace{-4mm}
\label{fig:td}
\end{figure*}

Let $\bm{X}=\{\bm{X}_1,\bm{X}_2, \ldots, \bm{X}_s\}$ be the set of samples in a dataset, where $s$ is the size of this set. Let $\bm{X}_{i}(j)$ be the $j$-th feature of $\bm{X}_i$. Given a black box classifier $\bm{y}_i = P(\bm{X}_i)$ of this dataset, the goal is to select the $n_e$ most important units of features of size $s_e$ from the input sample $\bm{X}_i$, that can be used to reconstruct a sample that achieves high prediction accuracy in the black box classifier. Fig.~\ref{fig:td} shows the DoRaR training framework and Algorithm.~\ref{alg:DoRaR} summarizes the training procedure.

\begin{algorithm}
 \scriptsize
  \caption{DoRaR Feature Selector Training Algorithm}  
  \label{alg:DoRaR}  
    \begin{algorithmic}
    \Require
    $\bm{X}\in R^{s\times m}$, training set samples, where $s$ and $m$ are dataset size and feature size;
    $\bm{E}\in R^{s\times m}$, reference background noise used to filling the non-selected features, where $\bm{E}_i(j)$, is randomly sampled from feature $j$ in the training dataset.
    $P$, classifier to be interpreted;
    \Ensure  
     $F_{n_e,s_e}(\cdot;\theta_F)$, feature selector that returns mask to select $n_e$ explanation units each consist of $s_e$ features given an input instance $\bm{X}$
    \renewcommand{\algorithmicrequire}{ \textbf{Select:}}
    \Require
    $d$, mini-batch size;
    $\lambda$, learning rate;
    $t$, training steps;
    $\alpha$, control parameter that balance selected and non-selected features;
    $\beta$, control parameter that balance prediction loss and reconstruction loss;
    $n_e$, number of output explanation unit;
    $s_e$, size of output explanation unit;
    \For{$1,...,t$}
        \State Randomly sample mini-batch of size $d$, $\bm{X}\in R^{d\times m}$
        \For{$i=1,...,d$}
            \State
            $\bm{M}_i=F(\bm{X}_i;\theta_F)$
            \State $\widehat{\bm{X}}_i=G_1(\bm{X}_i\cdot \bm{M}_i+\bm{E}_i\cdot(1-\bm{M}_i);\theta_{G_1})$
            \State $\tilde{\bm{X}_i}=G_2(\bm{X}_i\cdot (1-\bm{M}_i)+\bm{E}_i\cdot \bm{M}_i;\theta_{G_2})$
            \State
            $\bm{y}_i=P(\bm{X}_i)$
            \State
            $\widehat{\bm{y}}_i=P(\widehat{\bm{X}}_i)$
            \State
            $\tilde{\bm{y}_i}=P(\Tilde{\bm{X}_i})$
            \State
            $l_1=\sum_{i=1}^{d}\sum_{j=1}^{m}\frac{1}{m\cdot d}\left|\bm{X}_i(j)-\widehat{\bm{X}}_i(j) \right|$
            \State
            $l_2=\sum_{i=1}^{d}\sum_{j=1}^{m}\frac{1}{m\cdot d}\left|\bm{X}_i(j)-\tilde{\bm{X}_i}(j) \right|$
            \State
            $l_3=\frac{H_{\bm{y}}(\widehat{\bm{y}})}{d}=-\frac{\sum_{i=1}^{d}\bm{y}_i\log(\widehat{\bm{y}}_i)}{d}$
            \State
            $l_4=\frac{H_{\bm{y}}(\tilde{\bm{y}})}{d}=-\frac{\sum_{i=1}^{d}\bm{y}_i\log(\tilde{\bm{y}}_i)}{d}$
        \EndFor
        \State
        $\theta_F = Adam(L=(1-\alpha)(l_3+\beta l_1)-\alpha(l_4+\beta l_2),\lambda)$
        \State
        $\theta_{G_1} = Adam(L=l_3+\beta l_1,\lambda)$
        \State
        $\theta_{G_2} = Adam(L=l_4+\beta l_2,\lambda)$
    \EndFor
    \end{algorithmic}
\end{algorithm}

 The current form of feature selection is intractable because we choose top $n_e$ explanation units with the highest importance scores between 0 and 1 out of all optional units. This operation breaks the gradient calculation. In order to solve this, we use the generalized Gumbel-softmax trick  \cite{chen2018learning,jang2016categorical}. This is a commonly used technique to approximate a non-differentiable categorical subset sampling with differentiable Gumbel-softmax samples. By approximating the discrete mask $\bm{M}_i^*$ with $\bm{M}_i$ defined by $\bm{M}_i = F(\bm{X}_i;\theta_F),0\leq \bm{M}_i\leq1$, we have the continuous approximation to the argmax function that chooses top $n_e$ units with the highest scores. Then, we can use standard back propagation to calculate the gradient through the argmax function and train the feature selector properly. After the feature selector is well trained, we can switch back to the discrete mask $\bm{M}_i^*$ by setting the top $n_e$ output units of the feature selector to 1 and the rest output units to 0 during testing.

In the following subsections, we will explain how the artifacts problem and the EPITE problem are solved by different improvement techniques in our DoRaR method. The evaluation and comparison of each single improvement technique is presented in \ref{apd:evaluation1}.

\subsection{Dealing with the Artifacts Problem}

 For those feature attribution methods that evaluate the selected features directly in the pre-trained classifier, unwanted artifacts could be the major problem that needs to be solved. Several previous works \cite{dabkowski2017real,du2018towards} have proposed methods to reduce artifacts, such as using the blurred image as background noise, resizing the mask generated from a low resolution intermediate layer, or introducing smoothness regularization terms. But none of them solve the artifacts problem fundamentally because they still need to evaluate the explanation, where the non-selected features are replaced with ad-hoc values, on the pre-trained classifier directly.

 In our algorithm, given the selected features, we train a generative model to reconstruct an sample from the selected features with background noise, then feed the reconstructed sample into the pre-trained classifier. The assumption is that, for a well-trained generative model, a better mask should be able to select those features that can reconstruct an sample which achieves higher accuracy in the pre-trained classifier. 
 
\begin{figure}[h]
\centering
    \subfigure[The x-axis feature is chosen as the explanation and maintained, while the y-axis feature is removed and replaced by 0, it is easy to train a generative model to reconstruct samples that satisfy the classifier.]{\label{fig:art3}\includegraphics[width=0.45\columnwidth]{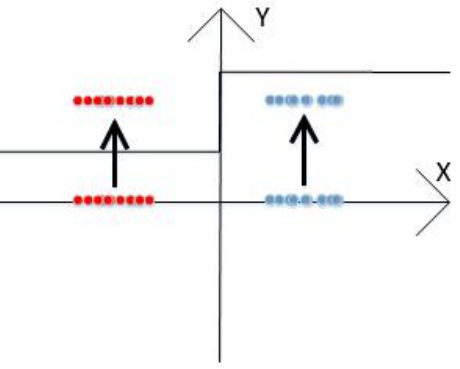}}
    \hspace{3mm}
\centering
    \subfigure[The y-axis feature is chosen as the explanation and maintained, while the x-axis feature is removed and replaced by 0, it is hard to train a generative model to reconstruct samples that satisfy the classifier.]{\label{fig:art4}\includegraphics[width=0.45\columnwidth]{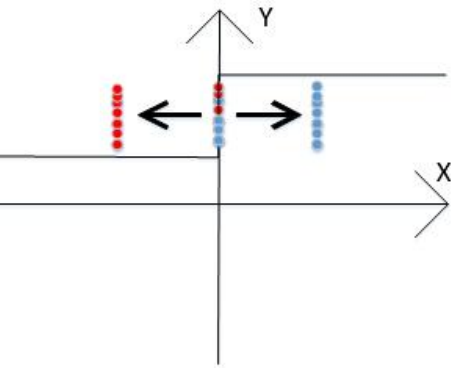}}
\centering
\caption{Two examples of choosing different features as the explanation. Arrows indicate the sample reconstruction through the generative model. By introducing the generative model, the difficulty of training a generative model for the example in the right figure to satisfy the classifier will let the feature selector prefer to choose the x-axis feature as the explanation.}

\label{fig:generativemodelsolveartifacts}
\end{figure}

 Fig.~\ref{fig:generativemodelsolveartifacts} shows an example that can verify our assumption. Following the example in Fig.~\ref{fig:artifacts}, it could appear that the y-axis feature is a better explanation for the classifier, as replacing the value of the y feature with zeros leads to a severe change in prediction accuracy (100\% to 50\%). However, with the generative model, if the x-axis value is kept, while the y-axis feature is replaced by some specific values, it is easy for the generative model to reconstruct the original data by learning to introduce a constant value of y to all data points. On the other hand, when choosing to keep the y-axis value and discard x, it is less likely to achieve high classification accuracy through a generative model -- using the y values only, a generator can no longer learn to reconstruct the original distinct classes.
 
\begin{figure}[h]

\centering\includegraphics[width=0.25\columnwidth]{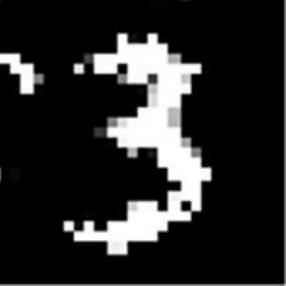}

\caption{An image generated from the selected 4 $4\times4$ pixels based chunks with the rest part filled with randomly selected value that follows the empirical distribution.}
\label{fig:fake image}
\end{figure}

 During the training process, we also introduced a reconstruction loss term for both feature selector and generative model to encourage the generated sample to be similar to the original sample. Fig.~\ref{fig:fake image} shows an example of an image generated from only 4 4x4 pixel based chunks of the original image. By doing this, the reconstructed sample will be closer to the natural data distribution.
 
\subsection{Dealing with the EPITE Problem}

For the EPITE problem, instead of using the commonly suggested Gaussian filter blurred input sample or average value  \cite{du2018towards} to fill in the non-selected area $1-\bm{M}$, we replace $\bm{X}(j)$ discarded by the mask by a noise sample $\bm{E(j)}$, randomly drawn from the training dataset, to minimize $I(\bm{X'};\bm{M})$.

\begin{figure}[h]

\centering
    \subfigure[Empirical distribution]{\label{fig:bn}\includegraphics[width=0.25\columnwidth]{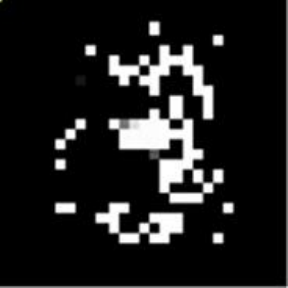}}
    \hspace{5mm}
\centering
    \subfigure[Gaussian distribution]{\label{fig:gn}\includegraphics[width=0.25\columnwidth]{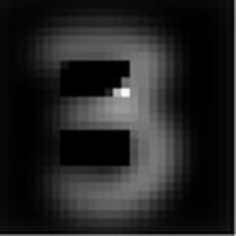}}
    \hspace{5mm}
\centering
    \subfigure[Mean value]{\label{fig:mn}\includegraphics[width=0.25\columnwidth]{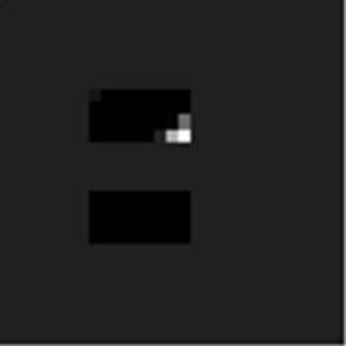}}
\centering

\caption{Selected 4 $4\times4$ pixels based chunks with the rest part filling with different types of background}

\label{fig:noised input}
\end{figure}

 Fig.~\ref{fig:noised input} shows examples where the area discarded by the mask is filled with pixel-wise random noise drawn from the training set, with the original discarded pixels blurred through a Gaussian filter, and with the pixel mean value over the training dataset. From the second and third images, we can easily find the position of selected explanation units. However, from the first image, it is much harder for the classifier to learn to locate the selected explanation units, then make corresponding predictions.

In practice, completely eliminating $I(\bm{X'};\bm{M})$ is impossible unless we use a perfect inpainter to fill the missing area, which introduces extra class information through the filling content. Therefore, in addition to apply proper background noise to minimizing $I(\bm{X'};\bm{M})$, we also reduce the impact of the EPITE problem by evaluating $I(\bm{\Bar{X'}};\bm{C})$, where $\bm{C}$ represents the class information, and $\bm{\Bar{X'}}=\bm{X}\cdot (1-\bm{M})+\bm{E}\cdot\bm{M}$.

For explanations with the EPITE problem, as depicted in Fig.~\ref{fig:EPITE}, it is important to note that the non-selected features still contain vital information and can achieve high prediction accuracy. To address this, our approach involves selecting explanations that demonstrate high accuracy when using the masked input (high $I(\bm{X'};\bm{C})$) and low accuracy when using the complementary masked input (low $I(\bm{\Bar{X'}};\bm{C})$). This selection process allows us to emphasize the class information present in the selected features ($I(\bm{X'};\bm{C}|\bm{M})$), rather than relying on the class information encoded in the mask learned by the retrained predictor ($I(\bm{X'};\bm{M};\bm{C})$). Because if the generative model $G_2$ can't recognize the complementary mask ($1-\bm{M}$) through $\bm{\Bar{X'}}$ then achieve high accuracy $a_2$ using the class information encoded in the mask, the generative model $G_1$ is less likely to take advantage from the class information encoded in the mask $M$ and has the EPITE problem either. For a detailed proof, please refer to \ref{apd:EPITE}.

%% file: evaluation_method.tex
\section{Evaluation scheme}
\label{sec:evaluation scheme}
Inappropriate feature attribution method evaluation can lead to unwanted result. For example, feature attribution method performance can be significantly impacted by evaluation parameters such as the order in which features are added or removed, specifically prioritizing the most relevant feature first (MoRF) or the least relevant feature first (LeRF). Different choices for these parameters may lead to conflicting outcomes. This inconsistency is unavoidable in some cases. For example, one single feature can independently improves classification accuracy a lot, like one pixel in a flag that indicates an existing color in certain area. It can be easily replaced by other features, like the surrounding pixels with the same color. Such feature can have high priority in adding the most relevant feature order, but there is also chances that it has high priority in removing the least relevant feature order when similar features are already exist. Further details on the limitations of choosing either the MoRF or LeRF order are described in Section~\ref{subsec:MoRF/LeRF}. Therefore, an appropriate evaluation metric is essential for feature attribution methods.

Similarly, the addition or removal of features one by one, based on a fixed feature attribution ranking, followed by the calculation of the corresponding accuracy, the inclusion area under curve (iAUC) and the exclusion area under curve (eAUC) can lead to unreliable results. This is because the contribution of a particular feature can be heavily influenced by the presence of other features. However, when calculating iAUC or eAUC, the contribution of a feature is only assessed in the presence of either more relevant or, respectively, less relevant features. We elaborate on this issue in Section~\ref{subsec:iAUC/eAUC}.

\subsection{Limitation with MoRF/LeRF order}
 \label{subsec:MoRF/LeRF}

The MoRF and LeRF are widely used orders for assessing the performance of feature attribution results. In LeRF, higher accuracy is preferred when removing least relevant features in order, while in MoRF, lower accuracy is preferred. However, \cite{rong2022consistent} identifies inconsistencies in the ranking of different attribution methods when different removal orders, such as MoRF/LeRF, are considered. The paper also proposes solutions to address this inconsistency. Nevertheless, we argue that these inconsistencies are inherent due to the mutual information between features.

\begin{figure}[htb]

\centering\includegraphics[width=0.75\columnwidth]{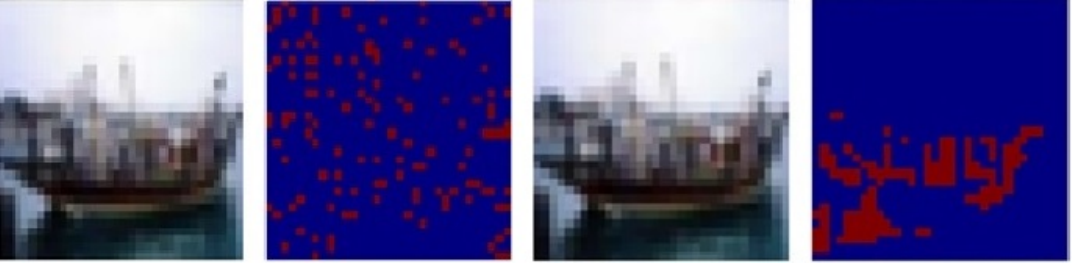}

\caption{Two examples of feature attribution result with high (left) and low (right) mutual information between features of high attribution score(red area) and features of low attribution score(blue area).}
\vspace{-4mm}
\label{fig:high mutual information}
\end{figure}

The left figure in Fig.~\ref{fig:high mutual information} illustrates a scenario where the selected features (marked in red) exhibit a scattered distribution with high attribution scores. In this case, regardless of whether we remove the most relevant or least relevant features first, the remaining features still contain valuable information and can yield high accuracy. Conversely, the right figure in Fig.~\ref{fig:high mutual information} presents a situation where the selected features cover important areas entirely, causing the remaining parts of the image to lack crucial information and resulting in low accuracy. 

In this case, if we remove features in the MoRF order, e.g. removing all selected features, it is possible for significant portions, such as the ship's outline, to be missed, leading to low accuracy and higher ranking for the right feature attribution result. On the contrary, if we remove features in the LeRF order, e.g. remove all non-selected features. The right figure will lose the whole sky area but the left figure will not miss any part entirely, which leads to higher accuracy and performance ranking for the left feature attribution result. Therefore, the inconsistent ranking of these two feature attribution results in the MoRF and LeRF orders is reasonably justified.

\subsection{Limitation with iAUC/eAUC metric}
 \label{subsec:iAUC/eAUC}
The metrics iAUC/eAUC are calculated based on a fixed feature attribution score vector. They determine the area under the accuracy curve derived by including or excluding features from the most/least relevant feature to the least/most relevant feature. However, assuming a universal feature attribution score is inappropriate. The attribution score of a feature also depends on the existence of other features.

As an example, consider Fig.~\ref{fig:high mutual information}. When we select only 2 pixels for explanation, any 2 pixels from the ship body cannot provide enough information to predict that the image is of a ship. However, 2 pixels from the background area, such as water and sky, can be much more informative.
\begin{equation}
    I(\{\bm{X_{sky},X_{water}}\};\bm{C})>I(\{\bm{X_{ship_1},X_{ship_2}}\};\bm{C})
\end{equation}
Here, $I(\{\bm{X_{sky},X_{water}}\};\bm{C})$ represents the mutual information between 2 pixels from background environment and the class variable. However, if we select 64 pixels instead of 2 pixels for explanation, selecting the ship can be more accurate in making the final classification, especially when 64 pixels are just enough to draw the rough outline of the ship.
\begin{equation}
\begin{aligned}
    &I(\{\bm{X_{sky},X_{water}}\};\bm{C}|\bm{X_{ship_3}...X_{ship_{64}}})\\
    <&I(\{\bm{X_{ship_1},X_{ship_2}}\};\bm{C}|\bm{X_{ship_3}...X_{ship_{64}}})
\end{aligned}
\end{equation}
While $I(\{\bm{X_{sky},X_{water}}\};\bm{C}|\bm{X_{ship_3}...X_{ship_{64}}})$ represents the mutual information between two pixels from the background and the class variable, given the other 62 pixels of the ship. Thus, the attribution score of a feature depends on the other features that have been selected. Evaluating the attribution score of a feature without limiting the existence of other features can lead to unreliable results. But in iAUC curve for example, each accuracy point in the curve only depends on the $n\ (1\leq n \leq m)$ most relevant features. Therefore, a more strict and precise control of other exist features is necessary in evaluation.

\subsection{A parameterized definition of evaluation metric}
As discussed in Section~\ref{subsec:MoRF/LeRF} and Section~\ref{subsec:iAUC/eAUC}, evaluating feature attribution result using single order or assuming a universal attribution score vector without considering interaction between features are unreliable. Therefore, we propose the following more reliable and more comprehensive definition for evaluating feature attribution method, so given a fixed number of selected features, all selected features are evaluated as a group, non-selected features are evaluated as a group either.

\textbf{Definition 1}(Feature Attribution Method). A $(n_e,s_e,a_1,a_2)$ feature selector of a black box classifier outputs $n_e$ explanation units, each of size $s_e$, such that:
\begin{enumerate}
    \item using only the $n_e$ units selected by the mask from the feature selector, it is possible to generate inputs to that black box, that are classified by the black box with accuracy at least $a_1$ relative to the target label, and
    \item using the complementary part of the masked inputs, excluding the selected $n_e$ units, it is possible to generate inputs to that black box, that are classified by the black box with accuracy at most $a_2$ relative to the target label.
\end{enumerate}

Under this definition, multiple types of feature selectors exist for a classifier to be explained. Users can specify either the format or the expected performance of the feature selector. With the same number and size of explanation units, one attribution method is better than another one with lower $a_1$ and higher $a_2$. In general, given two arbitrary feature attribution methods, we can use Definition 2 as a rule for comparing these two methods.

\textbf{Definition 2}(Feature Attribution Method Partial Order) Consider two feature attribution methods $F_1$ and $F_2$, with their format and performance defined by $({n_e}_1,{s_e}_1,{a_1}_1,{a_2}_1)$ and $({n_e}_2,{s_e}_2,{a_1}_2,{a_2}_2)$ respectively.
If ${a_1}_1 \geq {a_1}_2$ and ${a_2}_1 \leq {a_2}_2$ and ${n_e}_1 \leq {n_e}_2$ and ${s_e}_1 \leq {s_e}_2$
then $F_1$ is better than $F_2$ and we write $F_1\succeq F_2$.
If neither $F_1\succeq F_2$, nor $F_2\succeq F_1$, then $F_1$ and $F_2$ are \textbf{incomparable}.

Although this paper strictly adheres to the definition above, in the cases when ${n_e}_1 \leq {n_e}_2$ and ${s_e}_1 \leq {s_e}_2$, one could easily define other partial orders on the space of explanations, e.g., $F_1\succeq F_2$ if ${a_1}_1/{a_2}_1\geq{a_1}_2/{a_2}_2$ or if ${a_1}_1-w{a_2}_1\geq {a_1}_2-w{a_2}_2$ for some positive constant $w$.

\subsection{Our DoRaR evaluation strategy}

\begin{figure*}[htb]

\centering\includegraphics[width=1.0\linewidth]{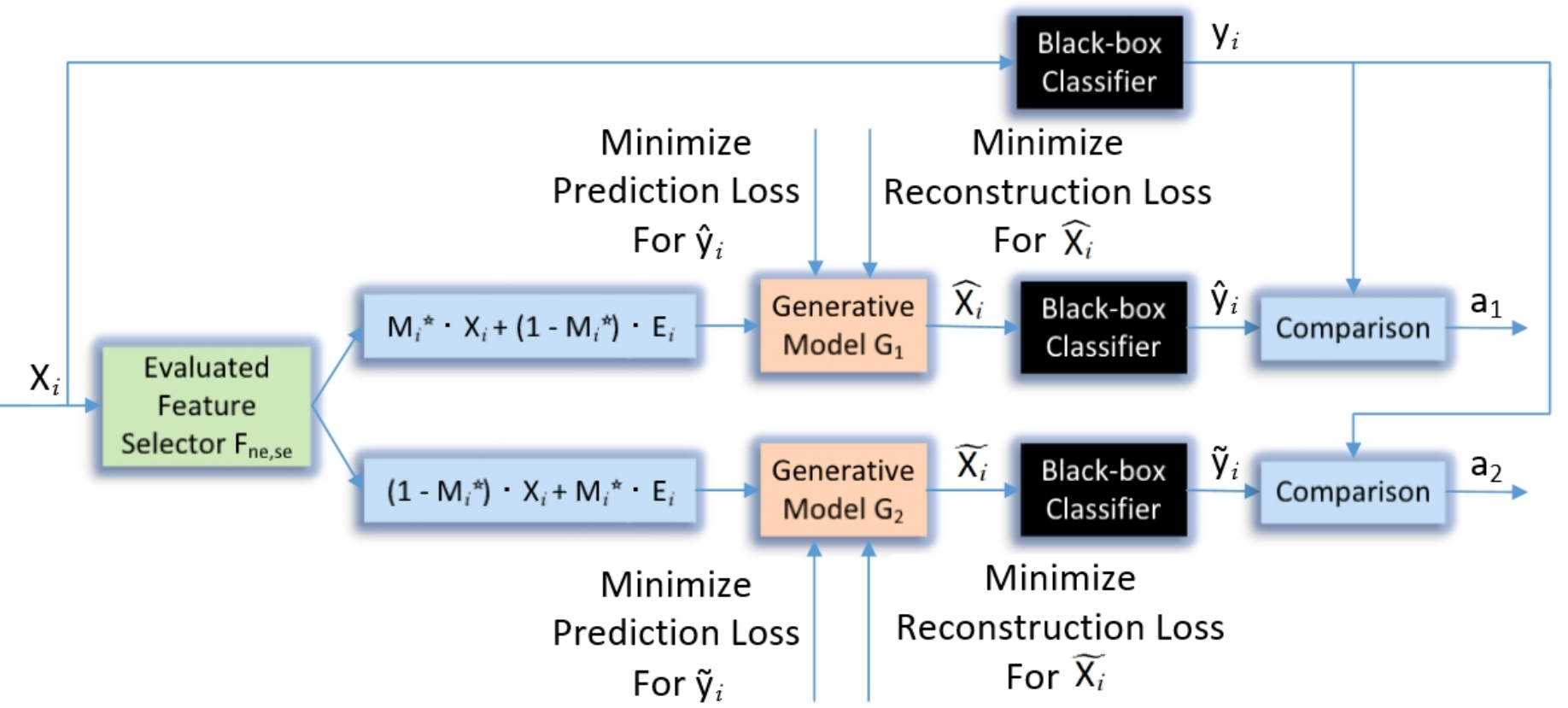}
\caption{Our DoRaR evaluation scheme. In evaluation, each feature selector based on corresponding feature attribution method outputs $n_e$ explanation units of size $s_e$. Their corresponding generative models are retrained based on prediction loss and reconstruction loss until converge.}
\label{fig:ts}
\end{figure*}

We use the scheme shown in Fig.~\ref{fig:ts} as our final evaluation strategy. It is used to test different feature attribution methods in the following section.

In this evaluation framework, the output format from the feature selector F is predefined, limiting feature interaction within the top $n_e\times s_e$ features. Adding prper background noise can reduce the performance of explanations suffering from the EPITE problem. Displaying the values of both $a_1$ and $a_2$ can also expose the EPITE problem through high $a_2$ values. Training generative models for each tested algorithm to evaluate selected features can prevent explanations that have the artifacts problem from receiving unfair advantages over those without the artifacts problem.

Additionally, for each feature selector being tested, it is worth mentioning that we utilize a simple structure for the generative models and train them using the same settings until converge. Consequently, any variations in evaluation results caused by perturbations in the generative models can be disregarded, the standard deviation caused by generative models is reported in the experiment result. This ensures that any observed differences in performance between feature selectors can be attributed to the selectors themselves, rather than the generative models used.

%% file: Comparison_to_other_methods.tex
\section{Comparison to other algorithms}
\label{evaluation2}

For comparison purposes, we implemented additional 6 algorithms: LIME, Grad, SmoothGrad, VIBI, Guided Feature Inversion, and Real-X, based on previous research \cite{ribeiro2016should,simonyan2013deep,smilkov2017smoothgrad,bang2021explaining,du2018towards,jethani2021have}. We evaluated these algorithms on MNIST, CIFAR-10, and a synthetic dataset based on user-mouse interaction behavior \cite{fu2022artificial} using testing scheme as Fig.~\ref{fig:ts}. All other comparison feature selectors were trained either directly on their authors' source code or using our own reproduction based on the pseudo-code in their paper. The major differences among the seven algorithms are summarized in Table \ref{table:algorithms differences}.

\begin{table*}[hbt]
\centering
\caption{Major differences between feature attribution methods.}
\label{table:algorithms differences}
\scalebox{0.9}{

\begin{tabular}{|c|c|c|c|c|l|}
\hline
\begin{tabular}[c]{@{}c@{}}Feature attribution\\ method\end{tabular} & \begin{tabular}[c]{@{}c@{}}Has Artifacts\\ Problem\end{tabular} & \begin{tabular}[c]{@{}c@{}}Has EPITE\\ Problem\end{tabular} & \begin{tabular}[c]{@{}c@{}}Generates different\\ explanations for\\ different models\end{tabular} & \begin{tabular}[c]{@{}c@{}}Needs access\\ to model\\ parameters\end{tabular} & \multicolumn{1}{c|}{\begin{tabular}[c]{@{}c@{}}Time Efficiency\\ (seconds/100 MNIST \\ image explanation )\end{tabular}} \\ \hline
LIME & Y & N & Y & Y & \multicolumn{1}{c|}{62.15} \\ \hline

Grad & Y & N & Y & Y & \multicolumn{1}{c|}{0.1953} \\ \hline

SmoothGrad & Y & N & Y & Y & \multicolumn{1}{c|}{29.09} \\ \hline
\begin{tabular}[c]{@{}c@{}}Guided Feature\\ Inversion\end{tabular} & Y & N & Y & Y & \multicolumn{1}{c|}{57.78} \\ \hline
Real-X & N & N & N & N & \multicolumn{1}{c|}{0.3306} \\ \hline
VIBI & N & Y & Y & N & \multicolumn{1}{c|}{0.3841} \\ \hline
DoRaR & N & N & Y & N & \multicolumn{1}{c|}{0.1707} \\ \hline
\end{tabular}
}
\vspace{-2mm}
\end{table*}

We test our DoRaR algorithm and other 6 feature attribution methods on the MNIST dataset, which includes 70000 28$\times$ 28 pixels images of hand written digits, 50000 for training, 10000 for validation and 10000 for testing respectively. Each feature attribution method is evaluated 5 times, mean and standard deviation are calculated as the result, so the perturbation caused by generative model can be seen from the result. A simple 2D CNN model is trained on the training set as the black box classifier to be explained which achieves 93\%  accuracy on the testing set.

We also test above 7 algorithms in the CIFAR-10 dataset, which consists of 60000 $32\times32$ color images in 10 classes, with 6000 images per class. Every feature selector is tested 5 times either. There are 50000 training images and 10000 testing images. A 2D CNN based model from the research \cite{yu2018deep} is trained as the black box classifier which achieves 95\% of test accuracy.

Since the MNIST and CIFAR-10 datasets lack ground truth explanations, apart from the principal analysis conducted earlier, we do not have concrete evidence to suggest that a feature attribution method with higher $a_1$ and lower $a_2$ in our evaluation scheme can effectively capture the true explanation. To address this limitation, we have created a synthetic dataset that incorporates a ground truth explanation. This dataset has been specifically designed to define the sole difference between the two classes of data, thereby allowing us to thoroughly test the performance of different feature attribution methods.

In the synthetic dataset, we collected mouse movement data from 18 users who performed a predefined task involving 10 consecutive movements in a specific pattern. This task was repeated a total of 2437 times. The mouse movement data was stored in the format of mouse cursor movement velocity in the x and y coordinates.

\begin{figure*}[htb]
 \vspace{-2mm}
\centering
    \subfigure[One of the mouse behavior samples chosen to be modified: before modification]{\label{fig:art3}\includegraphics[width=0.55\columnwidth]{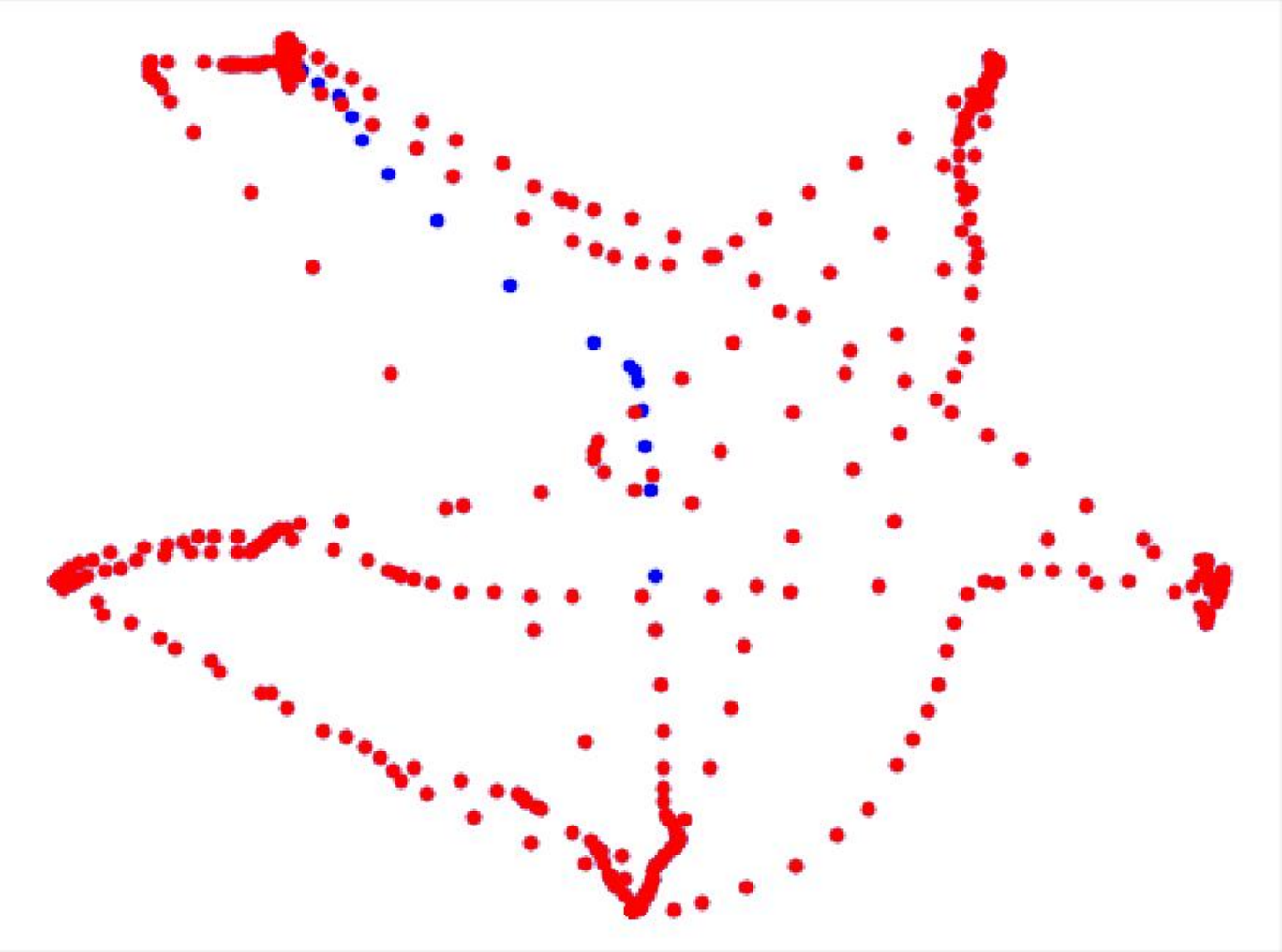}}
\hspace{3mm}
\centering
    \subfigure[One of the mouse behavior samples chosen to be modified: after modification]{\label{fig:art4}\includegraphics[width=0.55\columnwidth]{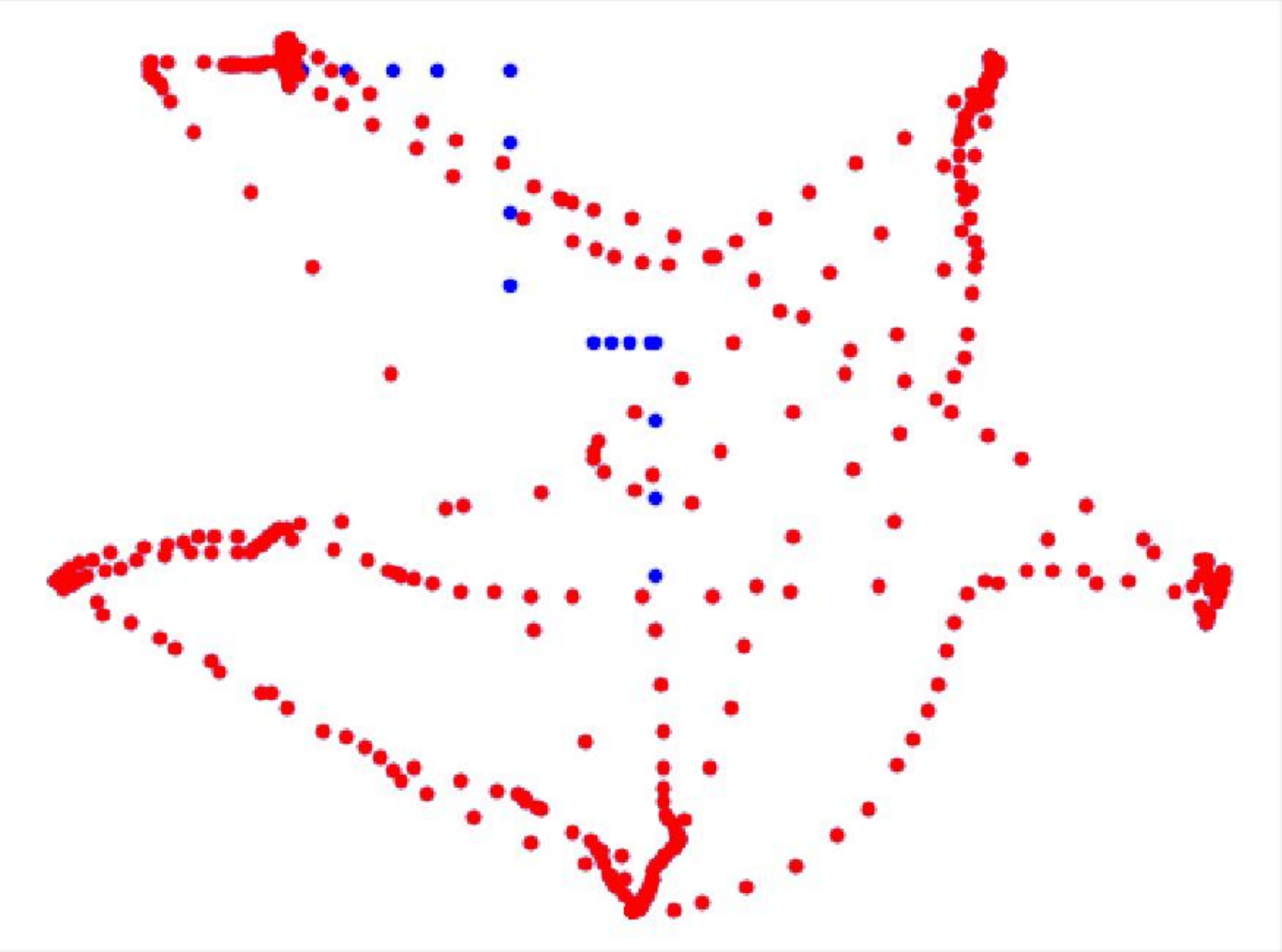}}
\centering
\caption{The synthetic dataset sample modification process involved modifying the blue points, which indicate the ground truth explanation, for the chosen half of the samples, while the remaining half of each user's samples were left unchanged.}
\vspace{-3mm}
\label{fig:synthetic dataset}
\end{figure*}

For each user, we modified half of their samples in the following manner: we replaced the first 16 points in the last movement with four consecutive horizontal or vertical segments. Each segment consisted of four points with equal movement speed, ensuring that the modified 16 points had the same start and end positions as the original movement. Figure~\ref{fig:synthetic dataset} provides a visual representation of how we modified a sample in the mouse behavior dataset. Please note that for convenience, these figures were plotted using x and y positions, while the classifier was trained using x and y velocities.

To evaluate the performance of different feature attribution methods, we used one-fifth of the modified and unmodified samples for testing purposes. Our pre-trained black-box classifier achieved an high accuracy of 99.78\% in correctly recognizing the modified samples in the testing set.

Our feature selector and generative model are built with 2D CNN and fully connected 3 layers neural networks respectively and are trained based on the black box classifier and the same training set. Details of the black box classifier and DoRaR architectures for MNIST dataset is same as described in \ref{apd:evaluation1}. See \ref{apd:cifar} for details of CIFAR-10 experiment. For the sequential data based synthetic dataset, 1D CNN is used for feature selector, \ref{apd:synthetic} shows more details of the mouse behavior based synthetic dataset experiment.

\subsection{Results of MNIST and CIFAR-10 Dataset}
\begin{table*}[hbt]
\centering
\caption{Mean and standard deviation of accuracy achieved through selected and non-selected features in MNIST dataset.}
\label{table:MNIST results}
\scalebox{0.9}{
\begin{tabular}{|c|cccccccc|}
\hline
Dataset & \multicolumn{8}{c|}{MNIST} \\ \hline
Explanation Format & \multicolumn{2}{c|}{4 Chunks} & \multicolumn{2}{c|}{8 Chunks} & \multicolumn{2}{c|}{12 Chunks} & \multicolumn{2}{c|}{24 Chunks} \\ \hline
Feature attribution method & \multicolumn{1}{c|}{\begin{tabular}[c]{@{}c@{}}Accuracy\\ ($a_1$/\%)\end{tabular}} & \multicolumn{1}{c|}{\begin{tabular}[c]{@{}c@{}}Accuracy\\ ($a_2$/\%)\end{tabular}} & \multicolumn{1}{c|}{\begin{tabular}[c]{@{}c@{}}Accuracy\\ ($a_1$/\%)\end{tabular}} & \multicolumn{1}{c|}{\begin{tabular}[c]{@{}c@{}}Accuracy\\ ($a_2$/\%)\end{tabular}} & \multicolumn{1}{c|}{\begin{tabular}[c]{@{}c@{}}Accuracy\\ ($a_1$/\%)\end{tabular}} & \multicolumn{1}{c|}{\begin{tabular}[c]{@{}c@{}}Accuracy\\ ($a_2$/\%)\end{tabular}} & \multicolumn{1}{c|}{\begin{tabular}[c]{@{}c@{}}Accuracy\\ ($a_1$/\%)\end{tabular}} & \begin{tabular}[c]{@{}c@{}}Accuracy\\ ($a_2$/\%)\end{tabular} \\ \hline
LIME & \multicolumn{1}{c|}{58.58$\pm$0.41} & \multicolumn{1}{c|}{89.78$\pm$0.29} & \multicolumn{1}{c|}{68.31$\pm$0.32} & \multicolumn{1}{c|}{87.95$\pm$0.45} & \multicolumn{1}{c|}{75.23$\pm$0.33} & \multicolumn{1}{c|}{86.48$\pm$0.29} & \multicolumn{1}{c|}{83.20$\pm$0.43} & 76.46$\pm$0.31 \\ \hline
Grad & \multicolumn{1}{c|}{18.07$\pm$0.33} & \multicolumn{1}{c|}{90.81$\pm$0.45} & \multicolumn{1}{c|}{24.86$\pm$0.37} & \multicolumn{1}{c|}{89.18$\pm$0.53} & \multicolumn{1}{c|}{42.45$\pm$0.57} & \multicolumn{1}{c|}{88.29$\pm$0.33} & \multicolumn{1}{c|}{75.31$\pm$0.44} & 78.13$\pm$0.24 \\ \hline
SmoothGrad & \multicolumn{1}{c|}{15.14$\pm$0.18} & \multicolumn{1}{c|}{90.82$\pm$0.39} & \multicolumn{1}{c|}{24.40$\pm$0.41} & \multicolumn{1}{c|}{89.87$\pm$0.36} & \multicolumn{1}{c|}{36.62$\pm$0.49} & \multicolumn{1}{c|}{88.33$\pm$0.41} & \multicolumn{1}{c|}{71.28$\pm$0.30} & 82.12$\pm$0.50 \\ \hline
Guided Feature Inversion & \multicolumn{1}{c|}{51.71$\pm$0.33} & \multicolumn{1}{c|}{89.05$\pm$0.28} & \multicolumn{1}{c|}{73.75$\pm$0.44} & \multicolumn{1}{c|}{85.23$\pm$0.33} & \multicolumn{1}{c|}{81.12$\pm$0.37} & \multicolumn{1}{c|}{79.98$\pm$0.30} & \multicolumn{1}{c|}{87.37$\pm$0.27} & 59.37$\pm$0.52 \\ \hline
Real-X & \multicolumn{1}{c|}{80.70$\pm$0.41} & \multicolumn{1}{c|}{88.80$\pm$0.37} & \multicolumn{1}{c|}{85.83$\pm$0.33} & \multicolumn{1}{c|}{85.60$\pm$0.31} & \multicolumn{1}{c|}{88.60$\pm$0.28} & \multicolumn{1}{c|}{82.40$\pm$0.29} & \multicolumn{1}{c|}{89.00$\pm$0.34} & 72.80$\pm$0.31 \\ \hline
VIBI & \multicolumn{1}{c|}{72.00$\pm$0.37} & \multicolumn{1}{c|}{88.03$\pm$0.57} & \multicolumn{1}{c|}{84.61$\pm$0.42} & \multicolumn{1}{c|}{85.80$\pm$0.35} & \multicolumn{1}{c|}{85.81$\pm$0.32} & \multicolumn{1}{c|}{80.70$\pm$0.41} & \multicolumn{1}{c|}{89.60$\pm$0.51} & 40.11$\pm$0.39 \\ \hline
DoRaR & \multicolumn{1}{c|}{81.30$\pm$0.44} & \multicolumn{1}{c|}{88.18$\pm$0.48} & \multicolumn{1}{c|}{86.29$\pm$0.45} & \multicolumn{1}{c|}{84.70$\pm$0.47} & \multicolumn{1}{c|}{88.65$\pm$0.34} & \multicolumn{1}{c|}{72.22$\pm$0.41} & \multicolumn{1}{c|}{89.10$\pm$0.52} & 13.03$\pm$0.46 \\ \hline
\end{tabular}
}
\end{table*}

\begin{table*}[hbt]
\centering
\caption{Mean and standard deviation of accuracy achieved through selected and non-selected features in CIFAR-10 dataset.}
\label{table:CIFAR10 result}
\scalebox{0.9}{
\begin{tabular}{|c|cccccccc|}
\hline
Dataset & \multicolumn{8}{c|}{CIFAR-10} \\ \hline
Explanation Format & \multicolumn{2}{c|}{4 Chunks} & \multicolumn{2}{c|}{8 Chunks} & \multicolumn{2}{c|}{64 Pixels} & \multicolumn{2}{c|}{128 Pixels} \\ \hline
Feature attribution method & \multicolumn{1}{c|}{\begin{tabular}[c]{@{}c@{}}Accuracy\\ ($a_1$/\%)\end{tabular}} & \multicolumn{1}{c|}{\begin{tabular}[c]{@{}c@{}}Accuracy\\ ($a_2$/\%)\end{tabular}} & \multicolumn{1}{c|}{\begin{tabular}[c]{@{}c@{}}Accuracy\\ ($a_1$/\%)\end{tabular}} & \multicolumn{1}{c|}{\begin{tabular}[c]{@{}c@{}}Accuracy\\ ($a_2$/\%)\end{tabular}} & \multicolumn{1}{c|}{\begin{tabular}[c]{@{}c@{}}Accuracy\\ ($a_1$/\%)\end{tabular}} & \multicolumn{1}{c|}{\begin{tabular}[c]{@{}c@{}}Accuracy\\ ($a_2$/\%)\end{tabular}} & \multicolumn{1}{c|}{\begin{tabular}[c]{@{}c@{}}Accuracy\\ ($a_1$/\%)\end{tabular}} & \begin{tabular}[c]{@{}c@{}}Accuracy\\ ($a_2$/\%)\end{tabular} \\ \hline
LIME & \multicolumn{1}{c|}{29.77$\pm$0.41} & \multicolumn{1}{c|}{51.05
$\pm$0.43} & \multicolumn{1}{c|}{31.22$\pm$0.54} & \multicolumn{1}{c|}{50.42$\pm$0.48} & \multicolumn{1}{c|}{32.95$\pm$0.40} & \multicolumn{1}{c|}{51.77$\pm$0.28} & \multicolumn{1}{c|}{34.97$\pm$0.49} & 49.54$\pm$0.38 \\ \hline
Grad & \multicolumn{1}{c|}{28.11$\pm$0.52} & \multicolumn{1}{c|}{51.57$\pm$0.32} & \multicolumn{1}{c|}{30.21$\pm$0.43} & \multicolumn{1}{c|}{50.80$\pm$0.48} & \multicolumn{1}{c|}{29.89$\pm$0.48} & \multicolumn{1}{c|}{52.29$\pm$0.50} & \multicolumn{1}{c|}{32.46$\pm$0.39} & 51.80$\pm$0.58 \\ \hline
SmoothGrad & \multicolumn{1}{c|}{29.85$\pm$0.38} & \multicolumn{1}{c|}{50.17$\pm$0.51} & \multicolumn{1}{c|}{32.52$\pm$0.42} & \multicolumn{1}{c|}{49.02$\pm$0.41} & \multicolumn{1}{c|}{31.74$\pm$0.49} & \multicolumn{1}{c|}{50.78$\pm$0.58} & \multicolumn{1}{c|}{33.05$\pm$0.57} & 49.17$\pm$0.44 \\ \hline
Guided Feature Inversion & \multicolumn{1}{c|}{30.36$\pm$0.55} & \multicolumn{1}{c|}{50.89$\pm$0.61} & \multicolumn{1}{c|}{35.83$\pm$0.62} & \multicolumn{1}{c|}{49.68$\pm$0.59} & \multicolumn{1}{c|}{25.16$\pm$0.51} & \multicolumn{1}{c|}{53.91$\pm$0.48} & \multicolumn{1}{c|}{31.52$\pm$0.61} & 51.29$\pm$0.56 \\ \hline
Real-X & \multicolumn{1}{c|}{32.80$\pm$0.50} & \multicolumn{1}{c|}{49.67$\pm$0.48} & \multicolumn{1}{c|}{39.89$\pm$0.48} & \multicolumn{1}{c|}{49.46$\pm$0.51} & \multicolumn{1}{c|}{38.11$\pm$0.53} & \multicolumn{1}{c|}{51.97$\pm$0.52} & \multicolumn{1}{c|}{44.32$\pm$0.43} & 50.83$\pm$0.50 \\ \hline
VIBI & \multicolumn{1}{c|}{26.60$\pm$0.39} & \multicolumn{1}{c|}{48.90$\pm$0.61} & \multicolumn{1}{c|}{30.34$\pm$0.48} & \multicolumn{1}{c|}{48.42$\pm$0.43} & \multicolumn{1}{c|}{29.24$\pm$0.33} & \multicolumn{1}{c|}{49.96$\pm$0.49} & \multicolumn{1}{c|}{35.15$\pm$0.42} & 49.46$\pm$0.61 \\ \hline
DoRaR & \multicolumn{1}{c|}{36.33$\pm$0.55} & \multicolumn{1}{c|}{49.11$\pm$0.44} & \multicolumn{1}{c|}{39.83$\pm$0.46} & \multicolumn{1}{c|}{48.84$\pm$0.39} & \multicolumn{1}{c|}{39.92$\pm$0.48} & \multicolumn{1}{c|}{49.47$\pm$0.47} & \multicolumn{1}{c|}{44.35$\pm$0.50} & 48.40$\pm$0.58 \\ \hline
\end{tabular}
}
\end{table*}

Table.~\ref{table:MNIST results} shows results of 7 feature attribution methods in scenarios of selecting 4, 8, 12 and 24 out of 49 $4\times 4$ pixels based chunks as explanations of the black box classifier. For the CIFAR-10 dataset, as shwon in Table.~\ref{table:CIFAR10 result} in addition to $4\times4$ chunk based explanations, we also test pixel based explanations. The reason for this is that important object characteristics in the CIFAR-10 dataset are typically smaller in size and distributed across a wider area of the image compared to the characteristics found in the MNIST dataset.

The result shows that in most scenarios our algorithm has higher $a_1$ and lower $a_2$ than all other algorithms. Real-X has comparable $a_1$ to our algorithm in some scenarios. But it has higher $a_2$ than our algorithm, especially in 12 chunks and 24 chunks scenarios of MNIST dataset as well as 64 pixels and 128 pixels scenarios of CIFAR-10 dataset.

\subsection{Result of Synthetic Dataset}
\label{subsec:Synthetic Dataset}

\begin{table*}[hbt]
\centering
\caption{Testing results in mouse behavior based synthetic dataset.}
\label{table:synthetic}
\scalebox{0.8}{
\begin{tabular}{|c|c|c|c|c|c|c|c|}
\hline
\begin{tabular}[c]{@{}c@{}}Feature\\ attribution\\ method\end{tabular} & LIME & Grad & \begin{tabular}[c]{@{}c@{}}Smooth\\ Grad\end{tabular} & \begin{tabular}[c]{@{}c@{}}Guided Feature\\ Inversion\end{tabular} & Real-X & VIBI & DoRaR \\ \hline
 \begin{tabular}[c]{@{}c@{}}Percentage of selected\\ features covering\\ ground truth explanation\end{tabular}& 21.01 & 32.70 & 79.20 & 13.31 & 25.43 & 0 & 76.51 \\ \hline
\end{tabular}
}
\end{table*}

Table \ref{table:synthetic} shows the results of 7 feature attribution methods in the scenario of selecting 4 segments, each consisting of 4 points, as explanations of the black box classifier. Since we have a ground truth explanation and our black-box classifier achieves very high accuracy, we can directly compare different feature attribution methods based on the hit rate of selected features that hit the ground truth explanation area.

The results show that our algorithm has a higher hit rate than all other algorithms except SmoothGrad. SmoothGrad has a comparable hit rate to our algorithm, but it has a much higher time cost. The VIBI algorithm achieves high prediction accuracy in their retrained predictor (97\%), but none of the selected features cover the ground truth explanation.

\subsection{Qualitative analysis}
\label{subsec:resultanalysis}

\begin{figure}[h]
\centering
    \subfigure[LIME Selects 4 Chunks]{\label{fig:lime_art}\includegraphics[width=0.9\columnwidth]{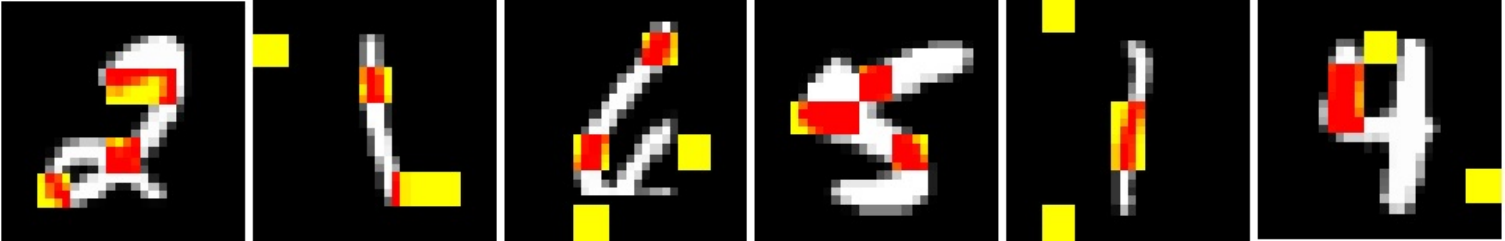}}
\centering
    \subfigure[Grad Selects 4 Chunks]{\label{fig:Grad_art}\includegraphics[width=0.9\columnwidth]{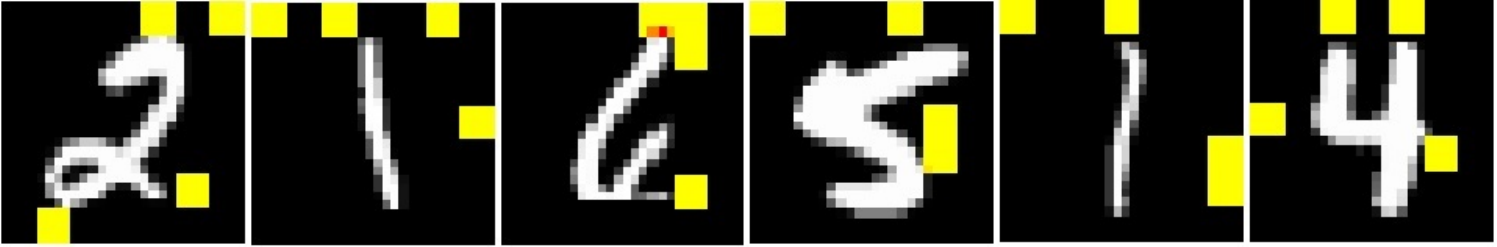}}
\centering
    \subfigure[SmoothGrad Selects 4 Chunks]{\label{fig:SG_art}\includegraphics[width=0.9\columnwidth]{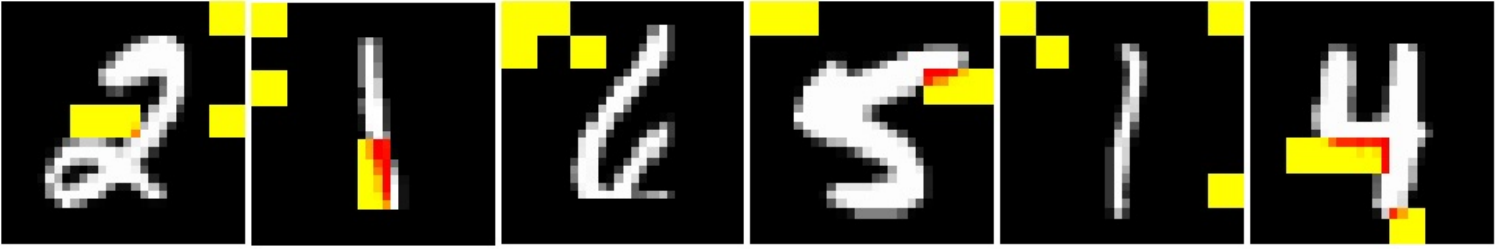}}
\centering
    \subfigure[DoRaR Based Feature Selector Selects 4 Chunks]{\label{fig:DoRaR_art}\includegraphics[width=0.9\columnwidth]{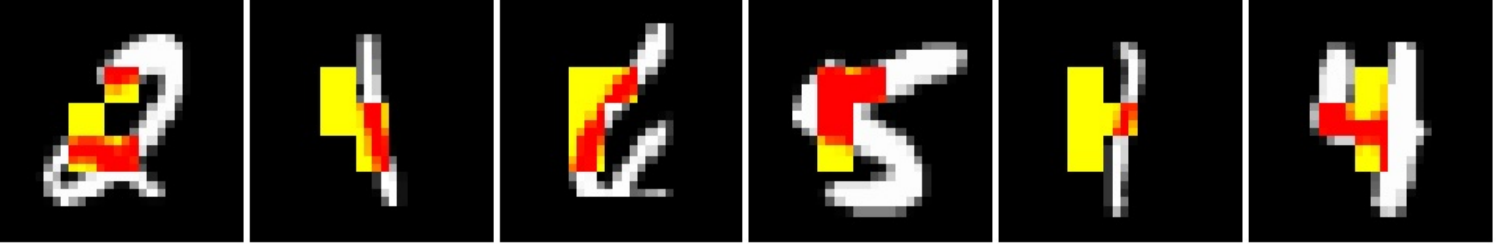}}
\centering
\caption{Different feature attribution methods select 4 $4\times 4$ pixels based chunks as explanation (marked in red if there is overlapping with the white digit and yellow otherwise). Examples show our algorithm can avoid the artifacts problem.}
\vspace{-3mm}
\label{fig:artifacts comparison}
\end{figure}

We choose some examples to illustrate the better performance of our algorithm than other algorithms. As depicted in Fig.~\ref{fig:artifacts comparison}, it is evident that feature attribution methods such as LIME, Grad, and SmoothGrad exhibit artifacts, particularly Grad and SmoothGrad. On the other hand, our DoRaR-based feature selector does not display such issues. This indicates that our approach effectively addresses the problem of artifacts that can arise with traditional feature attribution methods.

\begin{figure}[h]
\centering
    \subfigure[Guided Feature Inversion Selects 4 Chunks]{\label{fig:guided0}\includegraphics[width=0.9\columnwidth]{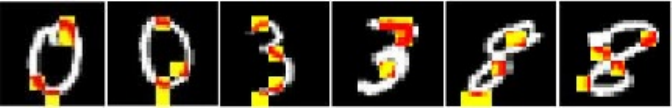}}
\centering
    \subfigure[DoRaR Based Feature Selector Selects 4 Chunks]{\label{fig:Ours0}\includegraphics[width=0.9\columnwidth]{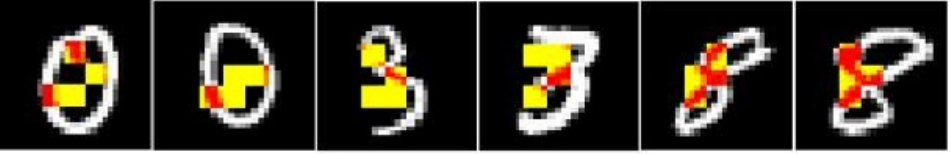}}
\centering
\caption{Examples show our algorithm can select those informative parts even though it has no overlapping with the digit while Guided Feature Inversion can not.}
\vspace{-3mm}
\label{fig:Guided vs Ours2}
\end{figure}

Fig.~\ref{fig:Guided vs Ours2} shows some examples of the better performance of our algorithm. Since selected chunks of the Guided Feature Inversion algorithm are based on weighted sum of activation values in different channels, those totally black areas, e.g. the central part of 0 and the left part of 3, which can differentiate 0 and 3 from 8, are less likely to be selected. On the contrary, our algorithm can select those informative parts even though they have no overlapping with the digit.

\begin{figure}[h]
\centering
    \subfigure[Real-X select 24 chunks]{\label{fig:RealX24}\includegraphics[width=0.9\columnwidth]{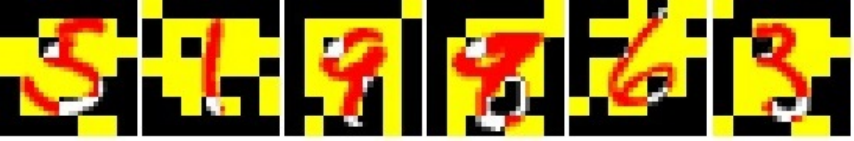}}
\centering
    \subfigure[DoRaR select 24 chunks]{\label{fig:Ours24}\includegraphics[width=0.9\columnwidth]{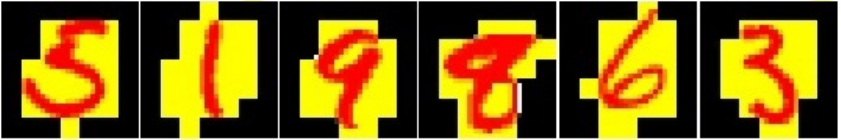}}
\centering
\caption{24 chunks with the highest probability scores: Real-X has similar $a_1$ to DoRaR based feature selector but its $a_2$ is higher. Therefore, the rest 25 chunks from Real-X are still very informative, while the DoRaR based feature selector selects those chunks that makes the rest part less informative.}
\vspace{-3mm}
\label{fig:RealX vs Ours}
\end{figure}

Fig.~\ref{fig:RealX vs Ours} shows the reason why Real-X has much higher $a_2$ in scenarios like selecting 24 chunks. It only optimizes $a_1$ in the training, so when the selected part contains enough information for classification, it is less likely to further polish the explanation. On the contrary, our algorithm will also minimize the information contained in the non-selected parts, which in turn improves the quality of the explanation.

\begin{figure}[h]
\centering
    \subfigure[VIBI Selects 4 Chunks]{\label{fig:guided0}\includegraphics[width=0.9\columnwidth]{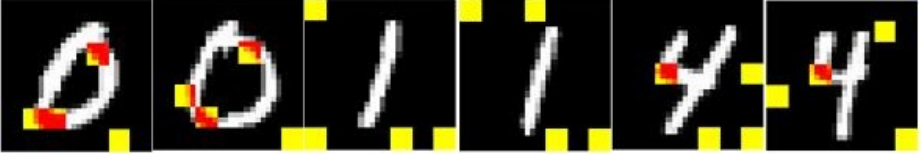}}
\centering
    \subfigure[DoRaR Base Feature Selector Selects 4 Chunks]{\label{fig:Ours0}\includegraphics[width=0.9\columnwidth]{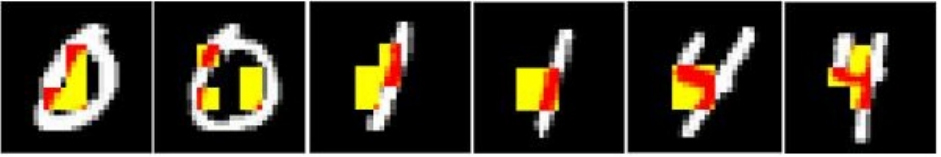}}
\centering
\caption{Examples of VIBI have the EPITE problem, while our algorithm does not have such a problem.}
\label{fig:VIBI vs Ours}
\end{figure}

Fig.~\ref{fig:VIBI vs Ours} shows some examples of the VIBI algorithm that encodes the prediction within the special location, e.g. uninformative corners, of masks. While they achieve high prediction accuracy in their jointly trained predictor, their prediction accuracy is lower in our testing scheme. Explanations selected by our algorithm don't show this problem.

\begin{figure}[h]
\centering
    \subfigure[DoRaR Selects 4 Chunks]{\label{fig:cifar10_background_1}\includegraphics[width=1\columnwidth]{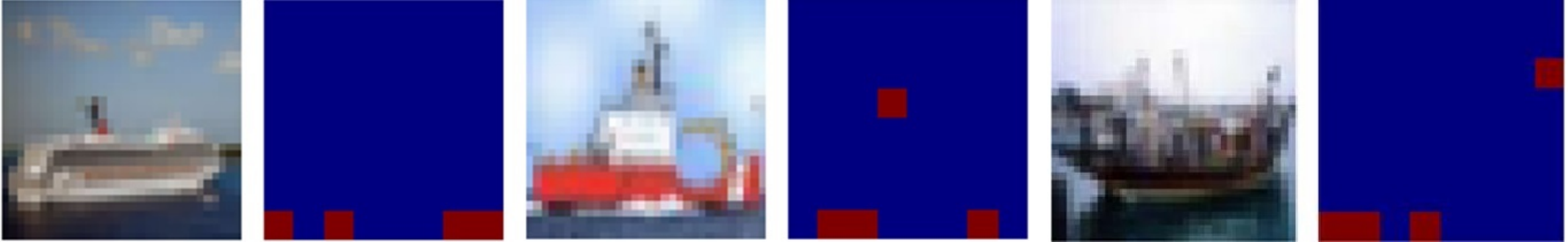}}
\centering
    \subfigure[DoRaR Selects 64 Pixels]{\label{fig:cifar10_background_2}\includegraphics[width=1\columnwidth]{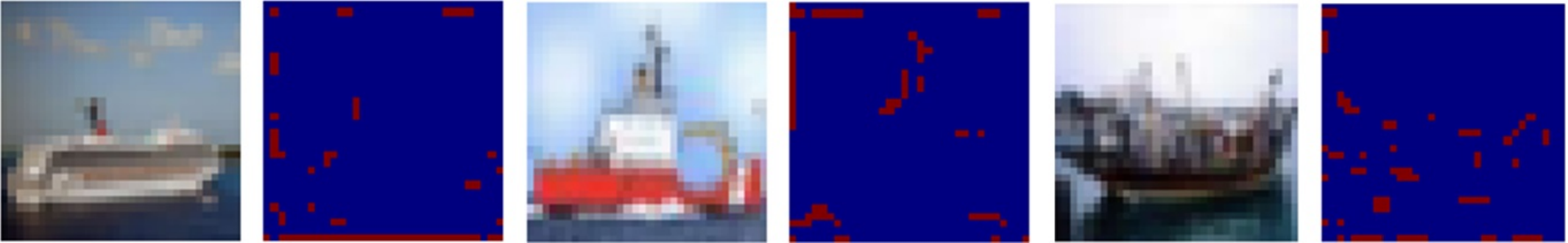}}
\centering
\caption{Examples of DoRaR selects 4 chunks and 64 pixels as model explanation of ship images.}

\label{fig:cifar10_background}
\end{figure}

For the CIFAR-10 dataset, except previous results, we have the following further findings. Like we discussed in section.~\ref{subsec:MoRF/LeRF} and section.~\ref{subsec:iAUC/eAUC}, for the pre-trained classifier, sometimes the background provides more information than the target itself for classification. For example, in Fig.~\ref{fig:cifar10_background_1}, if the limited number of selected chunks can't capture enough information of the ship's body, it prefer to select the background, e.g. sky and water.

Pixel based explanations can capture more details than chunk based explanations. For example, in Fig.~\ref{fig:cifar10_background_2}, while 4 $4\times4$ chunks based explanations focus on the background, pixel based explanations can capture both background information and some details of the ship body.

\begin{figure}[h]
\centering
    \subfigure[Real-X selects 128 pixels]{\label{fig:realx128}\includegraphics[width=0.65\columnwidth]{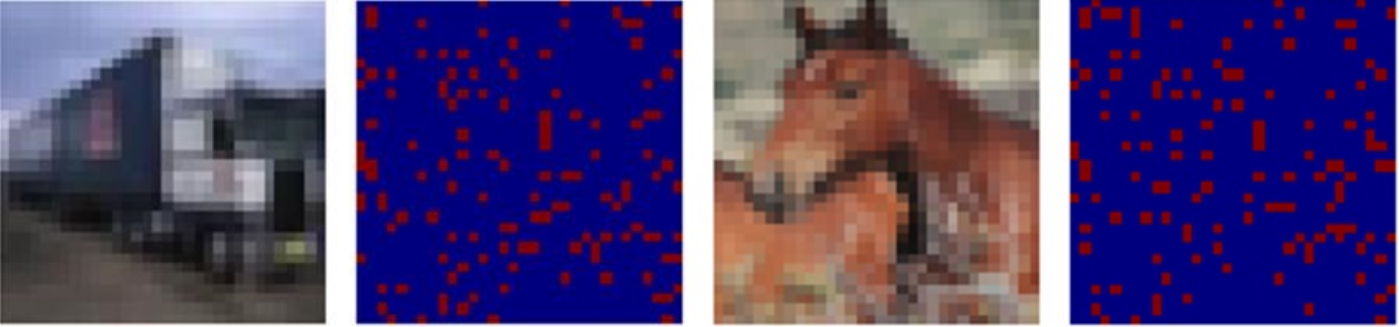}}
\centering
    \subfigure[DoRaR based feature attribution method selects 128 pixels]{\label{fig:Ours128}\includegraphics[width=0.65\columnwidth]{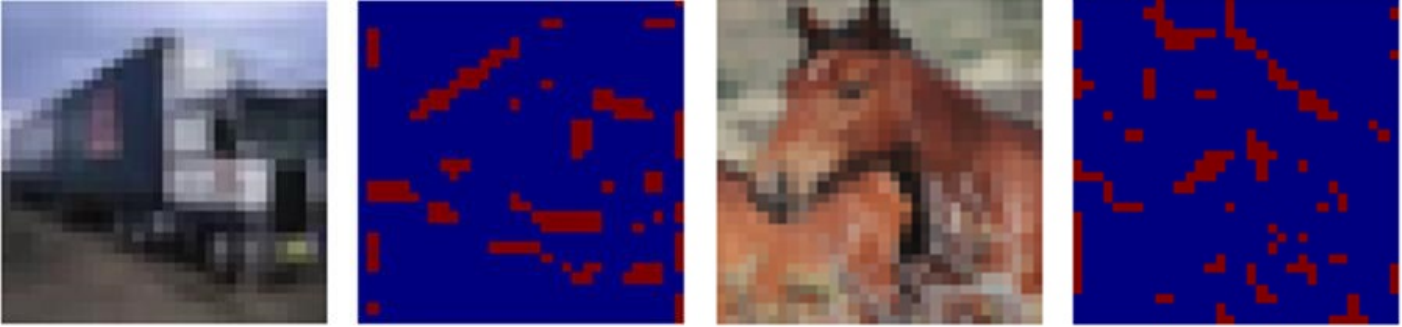}}
\centering
\caption{Examples of 128 pixels based model explanation from Real-X have a very scattered distribution, while the explanation from DoRaR based feature attribution method can highlight meaningful areas.}

\label{fig:cifar10 RealX vs Ours}
\end{figure}

However, pixels based explanations can be defective. Figure~\ref{fig:cifar10 RealX vs Ours} shows pixel-based explanations of the CIFAR-10 dataset from Real-X, which have a very scattered distribution. In contrast, our algorithm has a more concentrated feature selection that has much less mutual information with non-selected parts. This is caused by the term $I(\bm{X'};\bm{\Bar{X'}};{C}|\bm{M})$ in the target function \ref{equation:simplyfied target}. This example shows that when two feature attribution methods are incomparable under our partial order definition, we sometimes need to visually compare the specific explanations produced. Although we subjectively prefer the explanation of our algorithm, a more scattered feature selection as in Figure~\ref{fig:realx128} that allows a higher mutual information $I(\bm{X'};\bm{\Bar{X'}};{C}|\bm{M})$ can be generated by setting the training hyper-parameter $\alpha$ to a smaller value.

\begin{figure*}[h]
\centering
    \subfigure[LIME]{\label{fig:cifar10_lime}\includegraphics[width=0.35\columnwidth]{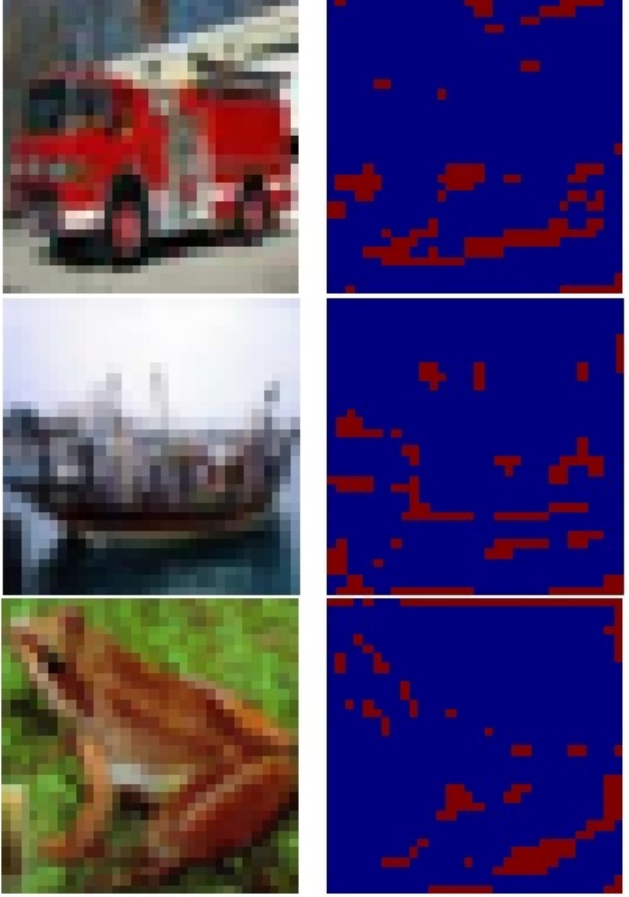}}
\centering
    \subfigure[Grad]{\label{fig:cifar10_Grad}\includegraphics[width=0.35\columnwidth]{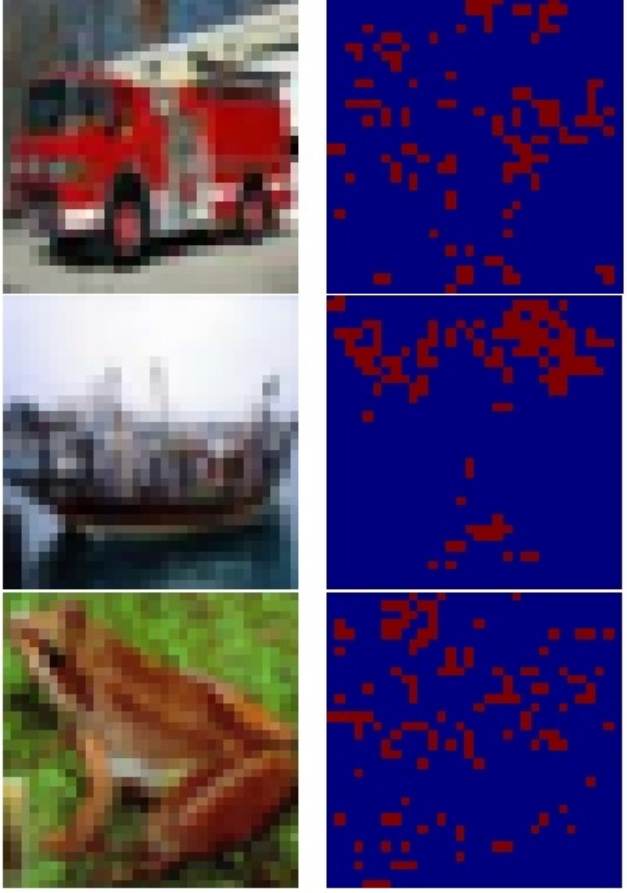}}
\centering
    \subfigure[SmoothGrad]{\label{fig:cifar10_SG}\includegraphics[width=0.35\columnwidth]{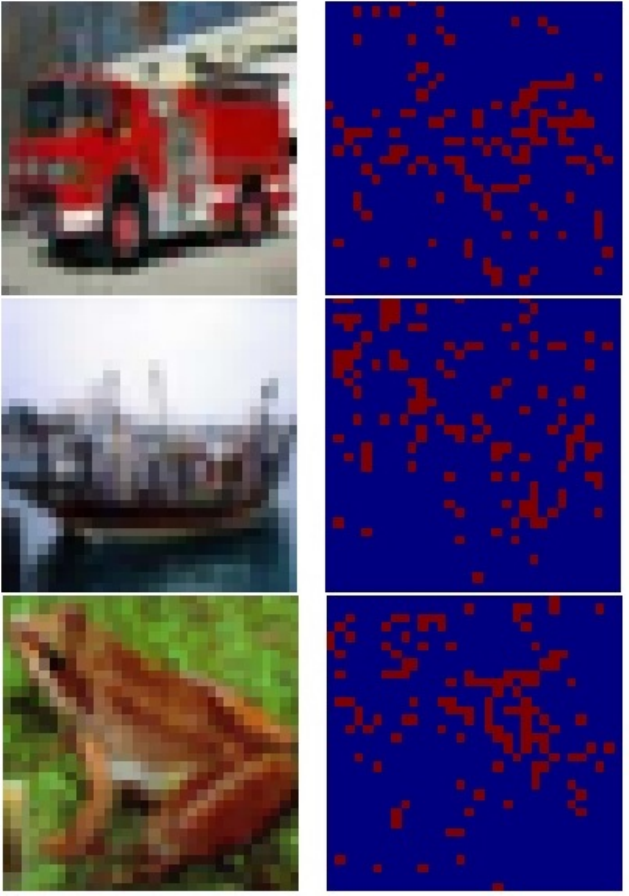}}
\centering
    \subfigure[Guided Feature Inversion]{\label{fig:cifar10_guided}\includegraphics[width=0.35\columnwidth]{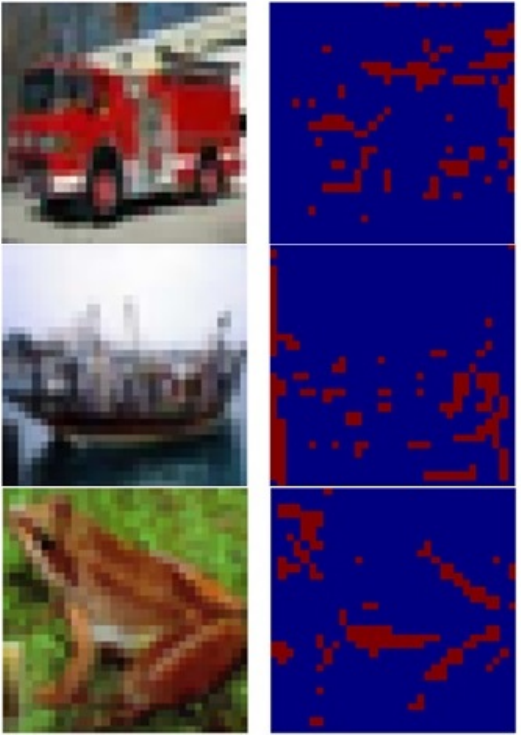}}
\centering
    \subfigure[Real-X]{\label{fig:cifar10_realx}\includegraphics[width=0.35\columnwidth]{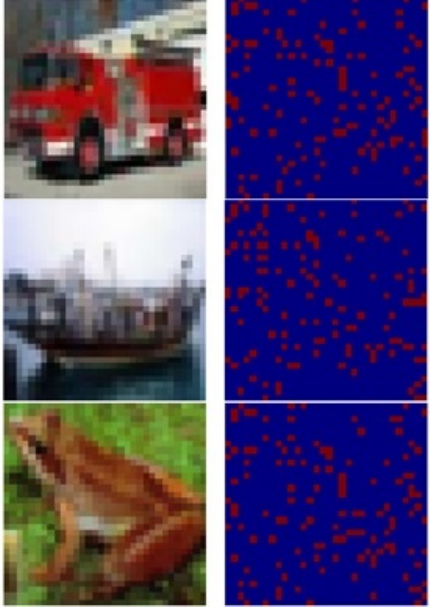}}
\centering
    \subfigure[VIBI]{\label{fig:cifar10_vibi}\includegraphics[width=0.35\columnwidth]{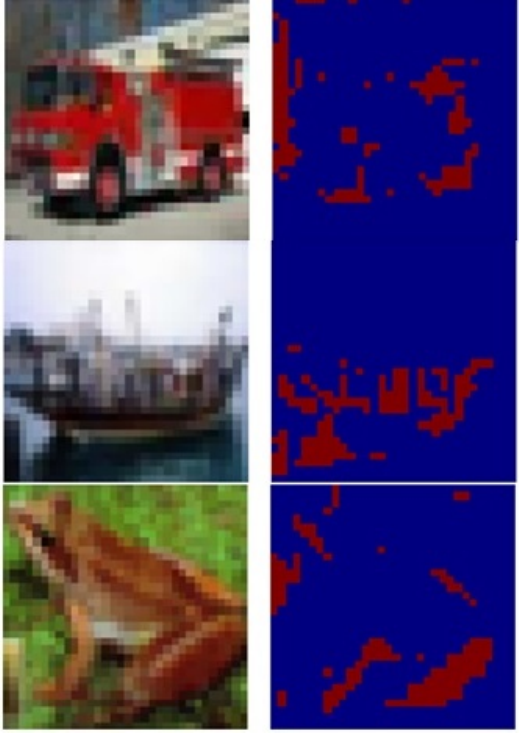}}
\centering
    \subfigure[DoRaR]{\label{fig:cifar10_dorar}\includegraphics[width=0.35\columnwidth]{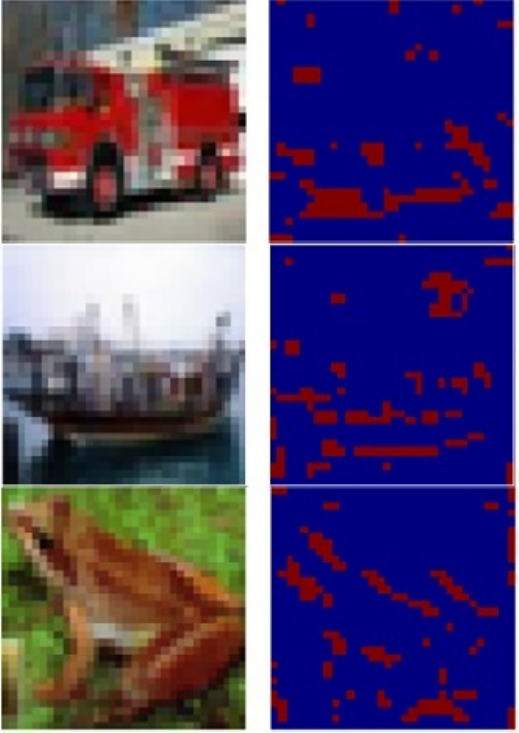}}
\centering
\caption{The CIFAR-10 dataset and explanations in format of 128 pixels provided by 7 feature attribution methods. The selected patches are colored red if selected as explanation and blue otherwise.}
\vspace{-3mm}
\label{fig:cifar10_7methods}
\end{figure*}

In the CIFAR-10 experiment, in addition to the previous results, we made further findings by observing Figure~\ref{fig:cifar10_lime} to Figure~\ref{fig:cifar10_dorar}. For example, in the truck image, our algorithm and LIME are the only two methods that focuses on the area between the tire and the ground, as well as the shadow in between. This is a distinctive pattern for truck images, which we did not expect before. But our method can find key features in real time while LIME can not.

%% file: discussion.tex
\section{Conclusion}
\label{sec:Discussion}
Based on our experimental observations, we conclude that minimizing prediction accuracy achieved by non-selected features can prevent the EPITE problem. If the mask that selects those features may cause information leakage, the complement mask has the same issue. Therefore, evaluating the performance of both selected and non-selected features can guarantee the reliability of explanations.

Reconstructing a new sample through a generative model to evaluate explanations is effective to solve the artifacts problem. Directly evaluating out of distribution masked input in the black-box classifier may lead to unexpected result, but a generative model that is trained to reconstruct the original input can help preventing such problem.

Adding the background noise in training can improve the performance of the explanation. The choice of background should minimize the mutual information between masked input and the mask. But the filling content should not provide extra class information.

It is crucial to predefine the size and number of basic explanation units in the evaluation process to ensure a fair comparison. However, the selection of these parameters should be customized to align with the specific requirements of the user. It is important to note that a feature attribution method that is effective in identifying important individual features may not perform as well when high attribution score features are evaluated as a group using our evaluation method, and vice versa. This highlights the importance of considering the specific context and goals when selecting an appropriate feature attribution method.

%% file: related_work.tex
\section{Other Related Work}
\label{subsec:related-work}

Various techniques have been developed to interpret machine learning models. Feature attribution methods that shows which feature contribute more to the classification can classified into two categories: real-time methods and non-real-time methods.

Non-real-time attribution methods require multiple iterations to learn an explanation for a single input sample. This category encompasses some perturbation-based methods \cite{zeiler2014visualizing, zhou2015predicting, zintgraf2017visualizing, fong2017interpretable, du2018towards}, some locally linear methods \cite{ribeiro2016should, lundberg2017unified} and gradient-based methods \cite{smilkov2017smoothgrad,sundararajan2017axiomatic,selvaraju2017grad} except the basic gradient-based method \cite{simonyan2013deep} which only require one backward propagation. These methods suffer from computational inefficiency, and the time required for certain techniques increases exponentially with the number of features. As a result, they are impractical for industry applications.

Real-time feature attribution methods, like those described in \cite{dabkowski2017real, bang2021explaining, jethani2021have, chen2018learning, yoon2018invase, dai2021towards, teso2019toward, kumar2021self}, require just one iteration to generate an explanation. This is typically accomplished by training a feature selector model. The feature selector model generates a feature selection mask for an input sample through one forward propagation.

Real-time feature attribution methods, despite their speed, face various challenges due to their inappropriate feature selector training process caused by the aforementioned problems. One such problem arises when feeding masked inputs directly to the black-box model for prediction loss calculation, which can introduce artifacts to the explanation.

Although retraining a new predictor might alleviate the artifacts issue \cite{hooker2018benchmark}, it can give rise to the EPITE problem. Some recent efforts has been made to overcome such problem, e.g. filling the non selected pixels with weighted mean of its neighbors \cite{rong2022consistent}, training feature selector and predictor separately \cite{jethani2021have}.

In addition, the authors of \cite{rong2022consistent} attempt to address the inconsistency in evaluation strategies based on different orders. However, a universal attribution vector that performs consistently in both MoRF and LeRF or iAUC and eAUC may not exist since feature interactions can significantly impact the feature attribution score.

%% file: Supplementary_Materials.tex
\section{Mitigating EPITE problem by evaluating prediction accuracy achieved by non-selected features}\label{apd:EPITE}

We demonstrate how we avoid EPITE problem when choosing explanation by evaluating the prediction accuracy achieved by non-selected features.

When measuring the prediction accuracy achieved through masked input, we expect $I(\bm{X'};\bm{C})$ to be high. On the contrary, we expect $I(\bm{\Bar{X'}};\bm{C})$ to be low, where $\bm{\Bar{X'}}=\bm{X}\cdot (1-\bm{M})+\bm{E}\cdot\bm{M}$ represents the features selected by complementary mask $1-M$ with the rest area filled with background noise $E$, and $\bm{C}$ represents the class variable. We can write:

\begin{equation}
    I(\bm{X'};\bm{C})=I(\bm{X'};\bm{C}|\bm{M})+I(\bm{X'};\bm{C};\bm{M}),
\end{equation}
\begin{equation}
    I(\bm{\Bar{X'}};\bm{C})=I(\bm{\Bar{X'}};\bm{C}|\bm{1-M})+I(\bm{\Bar{X'}};\bm{C};\bm{1-M})
\end{equation}
and we have
\begin{equation}
    H(\bm{M})=H(1-\bm{M}).
\end{equation}
Now, since we expect high prediction accuracy from the selected features and low prediction accuracy from the non-selected features, we can assume that we are actually evaluating
\begin{equation}
\begin{aligned}
&\,I(\bm{X'};\bm{C})-I(\bm{\Bar{X'}};\bm{C})\\
=&I(\bm{X'};\bm{C}|\bm{M})+I(\bm{X'};\bm{M};\bm{C})-I(\bm{\Bar{X'}};\bm{C}|\bm{M})\\
&-I(\bm{\Bar{X'}};\bm{M};\bm{C})
\end{aligned}
\label{equation:target}
\end{equation}

Since
\begin{equation}
\begin{aligned}
&I(\bm{\Bar{X'}};\bm{C}|\bm{M})\\
=&I(\bm{\Bar{X'}};\bm{C}|\bm{M},\bm{X'})+I(\bm{X'};\bm{\Bar{X'}};{C}|\bm{M})
\end{aligned}
\end{equation}
we can now write that 
\begin{equation}
\begin{aligned}
&I(\bm{X'};\bm{C}|\bm{M})-I(\bm{\Bar{X'}};\bm{C}|\bm{M})\\
= &I(\bm{X'};\bm{C}|\bm{M})-I(\bm{\Bar{X'}};\bm{C}|\bm{M},\bm{X'})\\
&-I(\bm{X'};\bm{\Bar{X'}};{C}|\bm{M})
\end{aligned}
\end{equation}
Using the chain rule for information, we have
\begin{equation}
\begin{aligned}
    &I(\bm{X'};\bm{C}|\bm{M})+I(\bm{\Bar{X'}};\bm{C}|\bm{M},\bm{X'})\\
    =&I(\bm{X'},\bm{\Bar{X'}};\bm{C}|\bm{M})\\
  \end{aligned}
\end{equation} 

Assuming the background noise $E$ contain no class information, we have
\begin{equation}
\begin{aligned}
    &I(\bm{X'},\bm{\Bar{X'}};\bm{C}|\bm{M})\\
    =&I(\bm{X};\bm{C}|\bm{M})\\
    =&t
  \end{aligned}
\end{equation} 

where $t$ is irrelevant to feature selection and treated as a constant here. Therefore, 

$I(\bm{\Bar{X'}};\bm{C}|\bm{M},\bm{X'})=t-I(\bm{X'};\bm{C}|\bm{M})$, s.t.
\begin{equation}
\begin{aligned}
    &I(\bm{X'};\bm{C}|\bm{M})-I(\bm{\Bar{X'}};\bm{C}|\bm{M})\\
    =&2I(\bm{X'};\bm{C}|\bm{M})-I(\bm{X'};\bm{\Bar{X'}};{C}|\bm{M})-t
\end{aligned}
\label{equation:9}
\end{equation}

Therefore, using function \ref{equation:9} our target function \ref{equation:target} can be simplified to
\begin{equation}
\begin{aligned}
    &I(\bm{X'};\bm{C}|\bm{M})+I(\bm{X'};\bm{M};\bm{C})-I(\bm{\Bar{X'}};\bm{C}|\bm{M})\\
    &-I(\bm{\Bar{X'}};\bm{M};\bm{C})\\
    =&2I(\bm{X'};\bm{C}|\bm{M})-I(\bm{X'};\bm{\Bar{X'}};{C}|\bm{M})\\
    &+I(\bm{X'};\bm{M};\bm{C})-I(\bm{\Bar{X'}};\bm{M};\bm{C})-t
\end{aligned}
\label{equation:simplyfied target}
\end{equation}

Using $I(\bm{X'};\bm{C})-I(\bm{\Bar{X'}};\bm{C})$ as the target function to evaluate, comparing with only evaluating $I(\bm{X'};\bm{C})$, we increase the weight of the class information from interested features $I(\bm{X'};\bm{C}|\bm{M})$ and reduce the weight of the source of EPITE problem $I(\bm{X'};\bm{M};\bm{C})$ largely in evaluation. Besides, if the EPITE problem gets severe, e.g. an improper background selection such as zero, both $I(\bm{\Bar{X'}};\bm{M};\bm{C})$ and $I(\bm{X'};\bm{M};\bm{C})$ will get close to the upper bound $H(C)$ so $I(\bm{X'};\bm{M};\bm{C})-I(\bm{\Bar{X'}};\bm{M};\bm{C})$ will get close to 0. Therefore, the effect of $I(\bm{X'};\bm{M};\bm{C})$ in target function will become more insignificant. The effect of minimizing $I(\bm{X'};\bm{\Bar{X'}};{C}|\bm{M})]$ is discussed in Section.~\ref{subsec:resultanalysis} through Fig.~\ref{fig:cifar10 RealX vs Ours}. In our DoRaR evaluation scheme, we compare both $a_1$ (prediction accuracy achieved by selected features) and $a_2$ (prediction accuracy achieved by non-selected features) without using a combined evaluation metric. In our feature attribution method proposed based on DoRaR, we use a hyperparameter $\alpha$ to balance the weight between $I(\bm{X'};\bm{C})$ (prediction loss achieved by selected features) and $I(\bm{\Bar{X'}};\bm{C})$ (prediction loss achieved by non-selected features).

\section{Evaluating Improvement of Background Noise (BN), Reconstruction Loss (RL) and Complementary Mask (CM) in MNIST dataset}\label{apd:evaluation1}

\subsection{Dataset and Experiment Setting}
\label{sec:Dataset and Experiment setting}

We tested each improvement technique, which included the inclusion of pixel-wise random background noise draw from empirical data distribution, the addition of a reconstruction loss term, the addition of a loss term corresponding to the complementary masked input, as well as their combinations by training a feature selector, then evaluate its performance in our evaluation scheme in the MNIST dataset. The dataset consists of 70,000 28$\times$28 pixel images of handwritten digits, with 50,000 images for training, 10,000 for validation, and 10,000 for testing. The experimental settings for the MNIST experiment in Section.~\ref{evaluation2} are the same as those described here.

\subsection{Model Structures}
\textbf{Black Box Classifier Structure.} For the black box classifier, we use a 2D CNN model which consists of 2 convolutional layers. The first convolutional layer has the kernel size 5 followed by a max-pooling layer with pool size 2 and a ReLU activation function. The second convolutional layer has the kernel size 2 followed by a 2D dropout layer and a ReLU function. Then it goes to a max-pooling layer with pool size 2 followed by a ReLU function. These two convolutional layers contain 10 and 20 filters, respectively. After these two convolutional layers, there are two fully connected layers with 20 and 10 units, respectively, connected by a ReLU function and a dropout layer in between. After that, there is a log-softmax calculation, so the final output returns a vector of log-probabilities for the ten digits.

\textbf{DoRaR Model Structure.}
Our DoRaR training procedure contains a feature selector and one or two generative models. The feature selector has 3 convolutional layers. The first two convolutional layers have kernel size 5 followed by a ReLU function and a max-pooling layer with pool size 2. The third convolutional layer has kernel size 1. Three convolutional layers have 16, 32 and 1 filters respectively. Then the output is flattened and the log-softmax value is calculated as the probability of choosing each optional explanation unit.

After up-sampling the input tensor from $7\times7$ to the standard image size $28\times28$, it is fed to the generative model, which consists of 2 fully connected layers, both with 784 units followed by a ReLU function and a Sigmoid function respectively.

\subsection{Parameter Selection}
\label{sec:Parameter Tuning}

\begin{figure}[H]
\centering\includegraphics[width=0.75\linewidth]{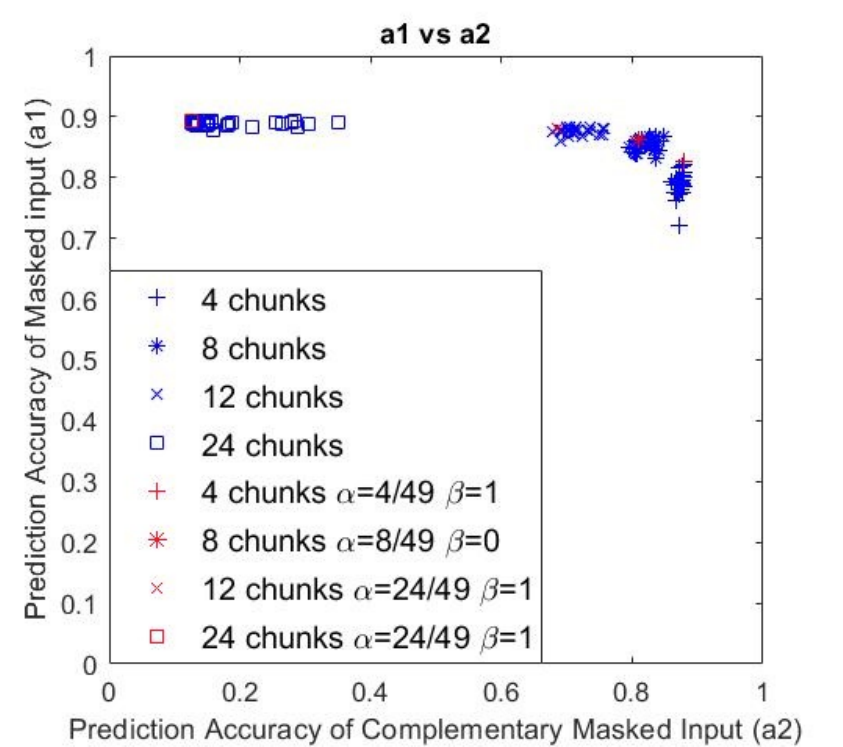}
\caption{Results of choosing different values for parameters, e.g., $\alpha=4/49$, $8/49$, $12/49$, $24/49$, $\beta=0$, $0.01$, $0.1$, $1$, $10$ in scenarios of selecting different number of chunks using our algorithm. Red marks represent parameters that achieve the best performance with relatively high $a_1$ and low $a_2$.}
\label{fig:tuning parameters}
\end{figure}

Values for parameters $\alpha$ and $\beta$ are determined by implementing each version of improvements (BN, RL, CM and all their combination), with different combination of parameter values. Fig.~\ref{fig:tuning parameters} shows examples of testing results of choosing different number of chunks with different values of $\alpha$ and $\beta$, e.g., $\alpha=\frac{4}{49}$, $\frac{8}{49}$, $\frac{12}{49}$, $\frac{24}{49}$ and $\beta=0$, $0.01$, $0.1$, $1$, $10$. Taking the 4 chunks scenario as an example, by choosing $\alpha = 4/49$ and $\beta=1$ (top red plus mark), we get the best prediction accuracy $a_1$ from selected chunks. In the following experiment, $\alpha$ is set to the fraction of the number of selected units to the total number of available units and $\beta$ is set to 1, which achieves relatively good overall performance in terms of both $a_1$ and $a_2$ as shown in red marks in Fig.~\ref{fig:tuning parameters}.

Other parameters are set as follows: learning rate for feature selector and generative models are $\lambda = 5e-4$. We use the Adam optimizer with batch size 100, while the coefficients used for computing running averages of gradient and its square are set as ($\beta1$ , $\beta2$ ) = (0.5, 0.999)

\subsection{Evaluated Methods}
\label{sec:Schemes and Metrics}
In order to study the effectiveness of each improvement method, we compare every single improvement method as shown in Fig.~\ref{fig:tested improvements} and all their combinations. We compare the performance of following the 8 schemes, in terms of the accuracy $a_1$ and the accuracy for the complementary masked input $a_2$:

\begin{figure}[h]
\centering
    \subfigure[Baseline]{\label{fig:baseline}\includegraphics[width=1\columnwidth]{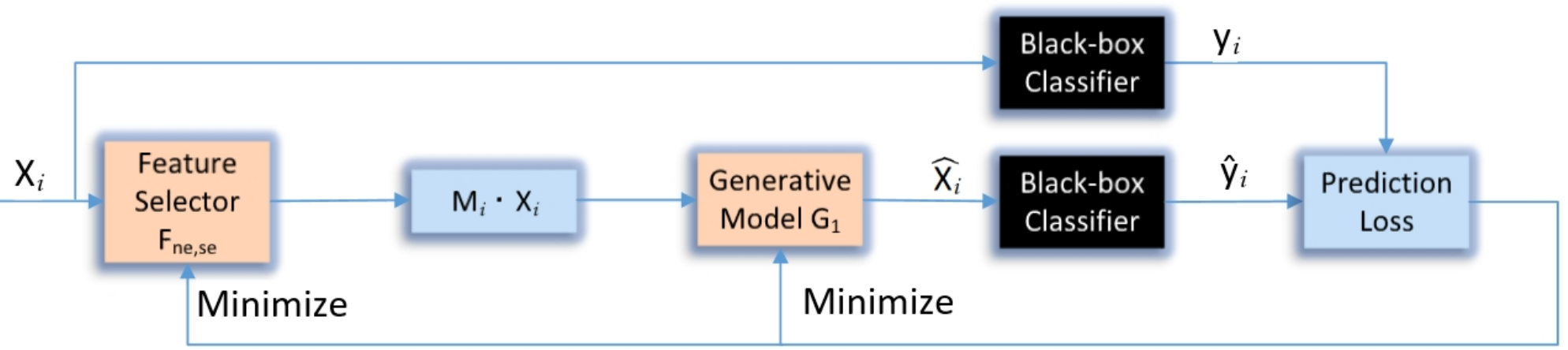}}
\centering
    \subfigure[Baseline + Background Noise]{\label{fig:baseline+noise}\includegraphics[width=1\columnwidth]{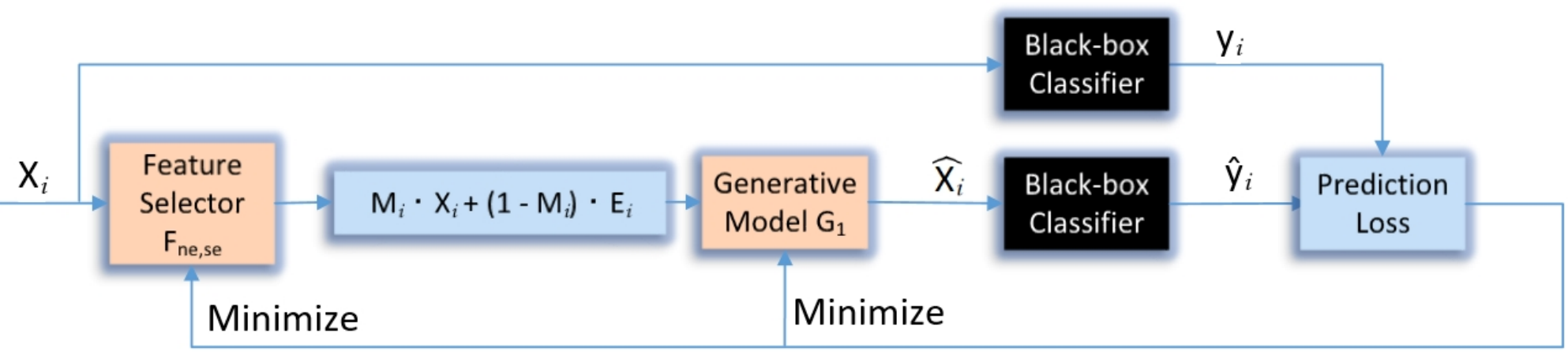}}
\centering
    \subfigure[Baseline + Reconstruction Loss]{\label{fig:baseline+reconstructionloss}\includegraphics[width=1\columnwidth]{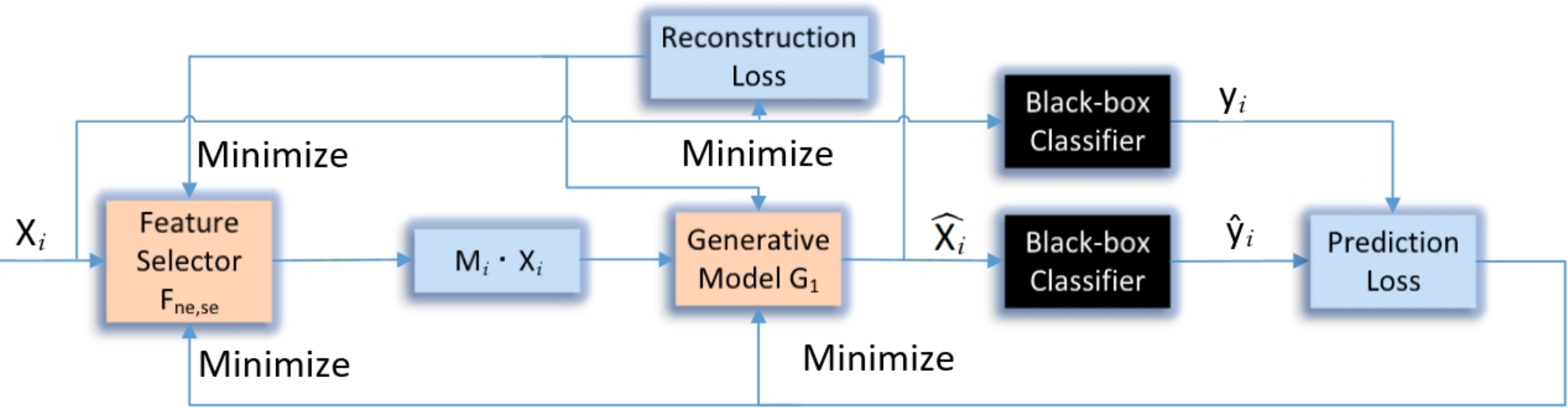}}
\centering
    \subfigure[Baseline + Complementary Masked input]{\label{fig:baseline+complementarymask}\includegraphics[width=1\columnwidth]{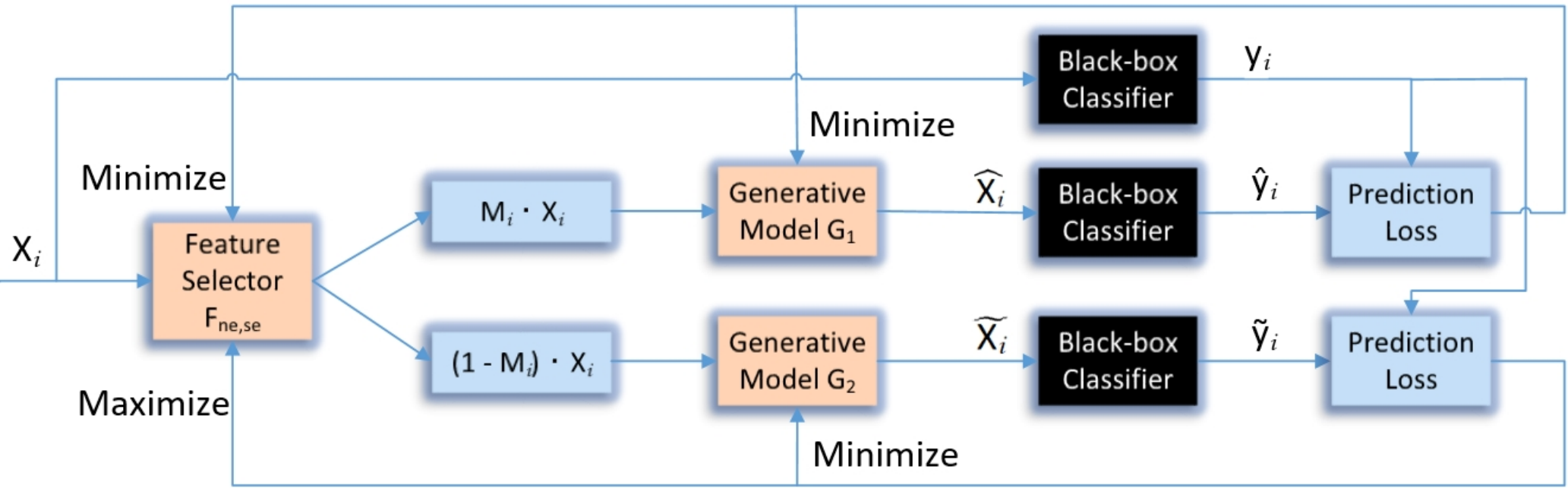}}
\centering
\caption{Diagrams of Improvement Methods}
\label{fig:tested improvements}
\end{figure}

\begin{enumerate}
    \item B: Baseline (Feature selector + Generative model without background noise, reconstruction loss and complementary masked input as shown in Fig.~\ref{fig:baseline}.)
    \item BN: Baseline + Background Noise
    \item RL: Baseline + Reconstruction Loss
    \item CM: Baseline + Complementary Masked Input
    \item BN+RL: Baseline + Background Noise + Reconstruction Loss
    \item BN+CM: Baseline + Background Noise + Complementary Masked Input
    \item RL+CM: Baseline + Reconstruction Loss + Complementary Masked Input
    \item BN+RL+CM (DoRaR): Baseline + Background Noise + Reconstruction Loss + Complementary Masked Input.
\end{enumerate}

\subsection{Results}
\label{subsec:self comparison results}
\begin{figure}[htb]
\centering\includegraphics[width=0.75\linewidth]{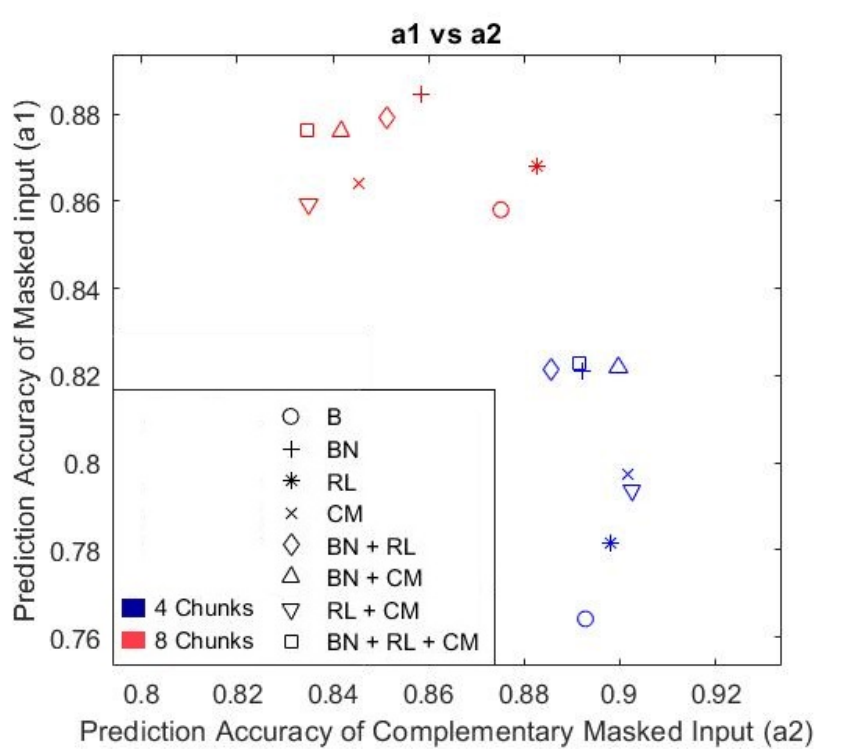}
\caption{Experiment testing results, B, BN, RL, CM indicating baseline and our improvements including background noise, reconstruction loss and complementary masked input. High $a_1$ and low $a_2$(top left) indicating better feature attribution method performance.}
\label{fig:sctr}
\end{figure}

Fig.~\ref{fig:sctr} shows the results of all possible combinations of the three improvement methods, for the 4 chunks and the 8 chunks scenarios. The results show that the baseline scheme always has the worst performance. The baseline combined with all three improvement methods has the best overall performance if we consider both $a_1$ and $a_2$. Single reconstruction loss has limited improvement, it has to be combined with other improvement methods. Given this result, we decide to use the algorithm that combines all three improvement methods as our DoRaR algorithm in this research.

\section{CIFAR-10 Dataset Experimental Settings}
\label{apd:cifar}
\subsection{Model Structures}
\textbf{Pre-trained Classifier Structure.} The pre-trained classifier is based on the model presented in the research \cite{yu2018deep}. It consist 3 basic convolutional layers followed by a batch normalization layer and a ReLU function then 4 layers of convolutional layer based tree blocks. All convolutional layers have the kernel size 3. Detailed structure of the convolutional layer based tree blocks is introduced clearly in \cite{yu2018deep}.

\textbf{DoRaR Structure.}
DoRaR model of the CIFAR-10 dataset is similar to that of the MNIST dataset with some small changes. For the feature selector, it has 3 convolutional layers. If the explanation unit is $4\times4$ chunk, then first two convolutional layers have kernel size 5 followed by a ReLU function. If it's pixel, then there is a max-pooling layer with the pool size 2 after each ReLU function. Third convolutional layer has kernel size 1. Three convolutional layers have 8, 16 and 1 filter respectively.

According to the image size of the CIFAR-10 dataset, the convolutional layers in generative model have $32\times32\times3$ units.

\subsection{Parameter Selection}
Parameter values for $\alpha$ and $\beta$ are tuned in the same way as described in S1. Other parameter values are same as the MNIST dataset except the learning rate for generative model which is set to $\lambda = 1e-4$.


\section{Synthetic Dataset Experimental Settings}\label{apd:synthetic}

\subsection{Model Structures}
\textbf{Black Box Classifier Structure.} The black box classifier consists of 1 2D convolutional layer with kernel size 2$\times$9 and 2 1D convolutional layers with kernel size 7 and 5, each followed by a ReLU function and a dropout layer in the front. These 3 CNN layers have 64, 96 and 128 filters respectively. First 2 CNN layers are followed by Max pooling layer with pool size 2 and Batch norm layer. After the CNN layers there is a bi-directory LSTM layer. Then it goes to two fully connected layers with 102400 and 128 units.

\textbf{DoRaR Structure.}
For the feature selector, it has 3 convolutional layers and 2 max-pooling layers with the same kernel size and filter number as in black-box classifier except the third convolutional layer that has only 1 output channel. According to the sample size of the mouse dataset, the linear generative model have 2 layers with 3200 and 3200 units.

\subsection{Parameter Selection}
Parameter values for $\alpha$ and $\beta$ are tuned in the same way as described in Section \ref{apd:evaluation1}. Other parameter values are same as the CIFAR-10 dataset except the batch size which is set to 50.